\documentclass[twoside,11pt]{article}

\usepackage{blindtext}

\usepackage[abbrvbib, preprint]{jmlr2e}

\let\proof\relax

\usepackage{mathrsfs, amsbsy, graphicx, amsmath, amssymb, algorithm, algorithmic, latexsym, amsthm}
\usepackage{xcolor}
\usepackage{subcaption}
\usepackage{graphicx}
\usepackage{booktabs}

\newtheorem{theo}{Theorem}
\newtheorem{prop}{Proposition}
\newtheorem{rem}{Remark}
\newtheorem{lem}[theorem]{Lemma}
\newtheorem{coro}[theorem]{Corollary}
\usepackage{placeins}

\theoremstyle{definition}

\usepackage{lastpage}

\ShortHeadings{Density-Guided Bootstrap Boundaries}{Gupta and Lee}
\firstpageno{1}

\begin{document}

\title{Density-Guided Bootstrap Boundaries for Latent World-Model Reinforcement Learning}

\author{\name Shashank Gupta \email shashang@umich.edu \\
       \addr Department of EECS\\
       University of Michigan\\
       Ann Arbor, MI 48109, USA
       \AND
       \name Pilhwa Lee \thanks{Corresponding author} \email pilhwa.lee@morgan.edu \\
       \addr Department of Mathematics\\
       Morgan State University\\
       Baltimore, MD 21251, USA
}


\maketitle

\begin{abstract}
Latent world models allow reinforcement learning agents to learn policies from imagined trajectories, yet determining the optimal depth of model reliance remains a fundamental challenge. Fixed imagination horizons apply uniform prediction depths regardless of trajectory reliability, exposing target values to compounding model error or over-relying on critic approximations. We introduce density-guided bootstrap boundaries, a trajectory-dependent mechanism that adapts the depth of imagined returns using a score-informed relative density proxy. The proxy accumulates the estimated divergence of a score-corrected auxiliary field formed by a feature displacement network and a noise-conditioned score estimator along imagined paths. We propose two gating mechanisms: a hard boundary that halts return expansion at a discrete state, and a soft boundary that continuously shifts the return recursion toward critic bootstrapping. Both variants preserve boundary value estimates and apply corresponding survival weights to actor and critic losses. Formulated independently of specific dynamics models and instantiated within DreamerV3, we prove that the resulting gated Bellman operator is a strict contraction mapping and that early bootstrapping monotonically reduces conditional target variance. We clarify the theoretical assumptions distinguishing our expansion proxy from exact path likelihoods. Across continuous-control tasks and Atari Pong, empirical evaluations show that dynamic boundary selection enhances sample efficiency and late-training performance over fixed-horizon and auxiliary-enabled baselines.
\end{abstract}


\section{Introduction}
\label{sec:introduction}

Learning a predictive model of the environment allows a reinforcement learning agent to improve its policy using imagined experience. Latent world models make this approach practical by representing observations in a compact state space and predicting the consequences of actions within that representation. DreamerV3 illustrates the effectiveness of combining a stochastic latent world model with actor--critic learning from imagined trajectories \citep{hafner2023dreamerv3}. These advances make model-based learning increasingly useful, but they leave an important question: how much of an imagined trajectory should contribute to a learning target before the agent relies on its value function? The depth of an imagined return determines a trade-off between model predictions and value estimates. Expanding a return further incorporates additional predicted rewards, but also exposes the target to errors in the learned dynamics and reward model. Bootstrapping earlier reduces this exposure while increasing dependence on the accuracy of the critic. A fixed maximum imagination horizon provides a simple way to control this trade-off. However, it applies the same limit to trajectories originating from different states and passing through different regions of the learned representation. Even when a return estimator mixes multiple prediction depths, a fixed mixing rule does not explicitly respond to a trajectory-dependent diagnostic of the model.

The question of when to trust a learned model has a substantial history. Model-Based Policy Optimization uses short model rollouts branched from real experience to limit the consequences of model error \citep{janner2019mbpo}. Stochastic Ensemble Value Expansion combines returns at different horizons using uncertainty estimates derived from ensembles \citep{buckman2018steve}. These approaches demonstrate that controlling model usage is an important design choice in its own right. Our work investigates a complementary mechanism: using a score-informed relative density proxy to select where an imagined return gives way to value bootstrapping.

We propose \emph{density-guided bootstrap boundaries} for latent world-model reinforcement learning. The framework associates each imagined state with a continuation gate that controls further return expansion. A hard gate selects a discrete bootstrap boundary, while a soft gate gradually shifts the return recursion toward the critic. In both cases, reaching a boundary preserves the value estimate at that state. The boundary therefore represents a decision about how much model-generated experience to use, rather than a prediction that the environment has terminated. Corresponding survival weights control the contribution of later imagined states to the actor and critic objectives. To construct the gating signal, we learn an auxiliary displacement field and a noise-conditioned score estimator from latent features inferred from replay observations. The score estimator supplies a correction to the auxiliary field. We estimate the divergence of the corrected field and accumulate its negative along imagined trajectories to obtain a relative density proxy. This construction gives the score estimator an explicit role in policy learning: its correction can change the accumulated signal, the selected bootstrap boundary, and consequently the targets and weights used by the actor and critic. The stochastic world model remains responsible for generating imagined states; the auxiliary field does not replace its transition distribution. The interpretation of this signal requires care. The accumulated quantity is an anchored density-change proxy associated with an auxiliary vector field, not an exact likelihood under the stochastic world model. In particular, imagined transitions are not generated by integrating that field, and estimating a score in a noise-smoothed feature space does not establish that low proxy values identify inaccurate predictions. We therefore treat the usefulness of the proxy for boundary selection as an empirical hypothesis. The relevant question is whether it supports better decisions about return expansion, rather than whether its magnitude alone can be interpreted as calibrated model uncertainty.

The bootstrap formulation is independent of a particular world-model architecture. Its use requires imagined trajectories, a value estimator, and a trajectory-dependent gate; our proposed density mechanism additionally requires suitable latent features and differentiable auxiliary models. We instantiate the method within DreamerV3, preserving its stochastic latent dynamics and using its actor--critic training procedure as the evaluation framework. This implementation allows us to study hard and soft boundaries within a common learning system. Our current implementation still generates the full imagined rollout, so adaptive return depth should not be interpreted as a demonstrated reduction in rollout computation.

A central evaluation issue is separating the information supplied by the density signal from the effects of shortening imagined returns. An improvement over a fixed maximum horizon could arise from a generally preferable shorter horizon, auxiliary training effects, or a change in the magnitude of the optimization objective. Accordingly, our evaluation framework distinguishes the unmodified agent, an auxiliary-enabled agent without boundary selection, and density-guided variants. Fixed-boundary controls, score ablations, and threshold sensitivity provide complementary tests of the proposed mechanism. Preliminary observations motivate these tests by showing that the score correction can substantially alter the divergence signal and that effective return depths vary across tasks and training stages; they do not by themselves establish that density-guided selection is superior.

Our contributions are threefold:
\begin{enumerate}
    \item We formulate hard and soft trajectory-dependent bootstrap boundaries for imagined actor--critic learning, preserving value bootstrapping and applying consistent survival weights to the learning objectives.
    \item We introduce a score-informed relative density proxy for boundary selection and specify the assumptions and limitations of its interpretation in a stochastic latent world model.
    \item We provide a DreamerV3 instantiation and an empirical
evaluation across continuous-control and visual discrete-action
tasks, including fixed-boundary controls, auxiliary-component
ablations, and diagnostics of effective imagination length.
\end{enumerate}
\section{Related Work}
\label{sec:related_work}

\paragraph{Latent world models and learning through imagination.}
Learning dynamics in a latent representation provides a way to use predictive models when observations are high-dimensional. PlaNet combines deterministic and stochastic latent transitions with online planning from image observations \citep{hafner2019planet}. Dreamer subsequently learns an actor and a value function through trajectories generated in latent space, propagating learning signals through imagined outcomes \citep{hafner2020dreamer}. DreamerV3 develops this approach into a broadly applicable world-model agent \citep{hafner2023dreamerv3}. Our work addresses how imagined experience enters the return estimator and training objective. The bootstrap-boundary formulation does not require a particular world-model architecture, although our implementation and empirical evidence are based on DreamerV3. We retain its stochastic transition model and introduce auxiliary models for selecting the depth of return expansion.

\paragraph{Model usage and adaptive value expansion.}
The benefits of model-generated experience depend on how prediction errors interact with policy and value learning. Model-Based Value Expansion (MVE) combines a finite sequence of model-predicted rewards with a terminal value estimate, using a fixed expansion depth \citep{feinberg2018mve}. Model-Based Policy Optimization (MBPO) uses short model rollouts branched from real experience to control the effects of model error during policy optimization \citep{janner2019mbpo}. These methods motivate treating the extent of model usage as a separate algorithmic choice. Stochastic Ensemble Value Expansion (STEVE) is particularly relevant: it constructs targets at multiple rollout depths and combines them using variance estimates from ensembles, adapting the mixture for individual examples \citep{buckman2018steve}. Consequently, adapting return depth to individual trajectories is not itself the novelty of our approach. Our proposed mechanism uses a score-informed relative density proxy to construct hard or soft continuation gates within the return recursion. These gates also induce survival weights on actor and critic losses. We investigate this density-based selection rule without interpreting it as an ensemble estimate of prediction uncertainty or asserting that it minimizes target variance.

\paragraph{Pessimism and restrictions on model-generated experience.}
Offline model-based reinforcement learning also studies how to limit the influence of unreliable predictions. MOPO penalizes predicted rewards using a measure of dynamics uncertainty \citep{yu2020mopo}. MOReL constructs a pessimistic model in which transitions classified as unknown lead to a penalized absorbing state \citep{kidambi2020morel}. These approaches address the risks of exploiting a learned model outside the support of a fixed dataset.

Our setting is online learning with a changing replay distribution, and our boundary has a different role. It ends or attenuates return expansion while retaining the critic estimate at the selected state. It does not impose an uncertainty penalty on rewards or declare the boundary to be an environment terminal state. This distinction matters because a bootstrap boundary transfers reliance from model predictions to the critic; it does not remove estimation error. Moreover, the proposed density proxy is not assumed to identify unknown transitions with the guarantees used in pessimistic offline formulations.

\paragraph{Adaptive imagination and computation allocation.}
Adaptive rollout decisions can also control planning computation. For example, RISE selects whether to continue imagination in world action models using estimates of risk, expected planning benefit, and computational cost \citep{lu2026rise}. This is related to our work at the level of making trajectory-dependent continuation decisions. However, our gates control the depth of actor--critic return expansion and the weights of imagined training samples. The current implementation generates the full imagined trajectory before applying these gates. Its effective return horizon therefore should not be equated with the amount of rollout computation performed.

\paragraph{Score estimation and probability-flow dynamics.}
Denoising score matching connects learning from noise-corrupted observations to estimating the gradient of a smoothed log density \citep{vincent2011connection}. Score-based generative modeling extends this connection to stochastic differential equations and associated probability-flow ordinary differential equations \citep{song2021scorebased}. These results motivate using learned scores to describe distributional structure and to construct corrections to vector fields.

In our method, a noise-conditioned estimator is trained on replay posterior features and contributes to an auxiliary vector field evaluated along imagined trajectories. The resulting signal is used for bootstrap selection. We do not use the score estimator to replace the stochastic world model with a diffusion sampler. In addition, the probability-flow interpretation requires assumptions about the relationship between the drift, diffusion, and score. Learning a displacement field and a score from replay features does not by itself establish those relationships for a discrete stochastic latent transition model. We therefore distinguish the continuous-time motivation from the operational density proxy used in our implementation.

\paragraph{Divergence-based density evolution.}
Continuous normalizing flows relate changes in log density to the divergence of the vector field generating a continuous trajectory. FFJORD uses stochastic trace estimation to make this computation scalable without forming the full Jacobian \citep{grathwohl2019ffjord}. We use the same computational principle to estimate the divergence of an auxiliary score-corrected field. The interpretation differs because our imagined trajectories are generated by the stochastic world model, rather than by integrating the auxiliary field. Accumulating estimated negative divergence along these trajectories yields an anchored relative proxy, not an exact world-model likelihood. The use of clipping and numerical trace estimation further affects the implemented signal. Our contribution concerns the use of this signal to control bootstrapping, rather than a new exact-likelihood model or a general guarantee of variance-free policy learning.

\paragraph{Position of the present work.}
The method brings together adaptive model usage, score estimation, and divergence-based diagnostics through a common bootstrap-gating formulation. The central empirical question is whether the density signal provides useful trajectory-dependent selection beyond the effects of a shorter return horizon or a change in loss scale. This motivates comparisons with fixed-boundary controls, auxiliary-enabled training without cutoff, and score ablations. Theoretical claims about the gated estimator must likewise account for critic error and the dependence of gates on sampled trajectories; properties of exact density transport do not automatically transfer to this implementation.
\section{Problem Formulation}
\label{sec:problem_formulation}

\subsection{Reinforcement Learning with a Latent World Model}
\label{subsec:latent_problem}

We consider a partially observed sequential decision process. At environment time $n$, the agent receives an observation $o_n$, selects an action $a_n$, and receives a reward $r_{n+1}$ after the transition. Let $\mathcal{H}_n=(o_0,a_0,r_1,o_1,\ldots,a_{n-1},r_n,o_n)$ denote its interaction history. The objective is to learn a policy that maximizes expected discounted reward,
\begin{equation}
J(\theta)
=
\mathbb{E}_{\pi_\theta,P}
\left[
\sum_{n=0}^{\infty}\gamma^n r_{n+1}
\right],
\label{eq:environment_objective}
\end{equation}
where $P$ denotes the environment dynamics, $\pi_\theta$ is the policy, and $\gamma\in[0,1)$ is the discount factor. Terminal states can equivalently be treated as absorbing states with zero subsequent rewards.

A world model with parameters $\phi$ represents the history through a latent state $z_n$ inferred from observations. We write this inference distribution as $q_\phi(z_n\mid\mathcal{H}_n)$ and the latent transition distribution as $p_\phi(z_{n+1}\mid z_n,a_n)$. The state $z_n$ includes any recurrent memory required by the model. A feature map $x_n=F_\phi(z_n)\in\mathbb{R}^{d}$ provides inputs to the actor, critic, reward predictor, and auxiliary models. This notation permits deterministic and stochastic components within $z_n$, as in recurrent latent world models \citep{hafner2019planet,hafner2020dreamer}. It does not require the feature distribution to have a density with respect to Lebesgue measure.

The policy is $\pi_\theta(a\mid x)$, and $V_\psi(x)$ denotes a scalar value estimate extracted from the critic. The world model also provides reward and continuation predictions. Replay data $\mathcal{D}$ contain observed interaction sequences used to train the world model and construct posterior latent features. The learned representation is an approximation to a sufficient state; we do not assume that it exactly preserves the underlying environment dynamics.

\subsection{Imagined Trajectories and Return Expansion}
\label{subsec:imagined_returns}

An imagined trajectory starts from a latent state inferred from replay and evolves under the policy and learned transition model:
\begin{equation}
a_t\sim\pi_\theta(\cdot\mid x_t),
\qquad
z_{t+1}\sim p_\phi(\cdot\mid z_t,a_t),
\qquad
x_t=F_\phi(z_t).
\label{eq:imagined_process}
\end{equation}
Here $t$ indexes imagination steps, separately from environment time $n$. We use $H$ to denote the number of transitions contributing to return expansion, so the trajectory contains states $x_0,\ldots,x_H$. This convention counts transitions rather than the number of feature entries in an implementation.

Let $\widehat r_{t+1}$ be the predicted reward associated with a transition and let
\begin{equation}
d_{t+1}=\gamma\widehat c_{t+1},
\qquad
\widehat c_{t+1}\in[0,1],
\label{eq:imagined_discount}
\end{equation}
where $\widehat c_{t+1}$ is the predicted probability of environment continuation. These quantities follow the reward and continuation indexing convention of the base world-model algorithm.

A finite-horizon $\lambda$-return is defined recursively by
\begin{equation}
G_t^\lambda
=
\widehat r_{t+1}
+
d_{t+1}
\left[
(1-\lambda)V_\psi(x_{t+1})
+
\lambda G_{t+1}^\lambda
\right],
\qquad
G_H^\lambda=V_\psi(x_H),
\label{eq:base_lambda_return}
\end{equation}
for $t=0,\ldots,H-1$ and $\lambda\in[0,1]$. The horizon bounds the maximum expansion depth, while $\lambda$ controls the mixture of intermediate bootstraps and longer returns. Our problem is to make this reliance on future imagined predictions depend on the trajectory.

Specifically, we seek a continuation gate $g_t\in[0,1]$ computed from the imagined prefix and learned auxiliary information. The gate is conceptually distinct from $\widehat c_t$: environment continuation describes whether the task continues, whereas $g_t$ describes whether the learning target should continue expanding through model predictions. A zero gate at a boundary should preserve the incoming reward and the discounted value estimate at that state. A unit gate should recover the base return recursion locally. The construction of the gates, their insertion into the recursion, and their associated loss weights are specified in Section~\ref{sec:method}.

\subsection{Score Estimation in Latent Feature Space}
\label{subsec:score_formulation}

Let $\mu_\phi$ denote the distribution of posterior features obtained from replay under the current world model. Because these features may contain discrete components or occupy a lower-dimensional subset of $\mathbb{R}^{d}$, we define score estimation through Gaussian smoothing. For a noise scale $\alpha>0$,
\begin{equation}
x\sim\mu_\phi,
\qquad
\epsilon\sim\mathcal{N}(0,I_d),
\qquad
\widetilde x=x+\alpha\epsilon.
\label{eq:feature_corruption}
\end{equation}
The perturbed distribution has density
\begin{equation}
p_{\phi,\alpha}(\widetilde x)
=
\int
\mathcal{N}(\widetilde x;x,\alpha^2 I_d)\,
\mu_\phi(dx),
\label{eq:smoothed_density}
\end{equation}
and its score is
\begin{equation}
s^\star_{\phi,\alpha}(\widetilde x)
=
\nabla_{\widetilde x}\log p_{\phi,\alpha}(\widetilde x).
\label{eq:score_definition}
\end{equation}

A noise-conditioned predictor $\epsilon_\eta(\widetilde x,\alpha)$ is trained using the denoising objective
\begin{equation}
\mathcal{L}_{\mathrm{noise}}(\eta)
=
\mathbb{E}_{x,\alpha,\epsilon}
\left[
\left\|
\epsilon_\eta(x+\alpha\epsilon,\alpha)-\epsilon
\right\|_2^2
\right].
\label{eq:noise_prediction_loss}
\end{equation}
The expectation over $\alpha$ follows a chosen positive noise-scale distribution. At the population optimum, for a fixed $\alpha$, the predictor estimates the conditional mean of the injected noise, giving the score parameterization
\begin{equation}
s_\eta(\widetilde x,\alpha)
=
-\frac{\epsilon_\eta(\widetilde x,\alpha)}{\alpha}.
\label{eq:noise_score_relation}
\end{equation}
This relation follows the denoising score-matching construction \citep{vincent2011connection,song2021scorebased}. A finite learned predictor only approximates the population score. Moreover, $\mu_\phi$ changes as replay and the world model evolve. The resulting score describes a smoothed replay-feature distribution; it is not automatically the score of the policy's imagined-state distribution or of the world-model transition kernel.

\subsection{Divergence as an Empirical Feature-Space Expansion Diagnostic}
In continuous dynamics governed by a vector field $v(x,u)$, local volumetric changes in feature space are tracked by the spatial divergence operator $\nabla_x \cdot v(x,u) = \text{Tr}(J_v(x))$. In latent world-model reinforcement learning, imagined state transitions follow a discrete stochastic process $z_{t+1} \sim p_\phi(z_{t+1} \mid z_t, a_t)$ rather than continuous integration along trajectories of $v(x,u)$.

Consequently, the spatial divergence of an auxiliary score-corrected vector field does not yield an exact path log-likelihood under the stochastic world model. Instead, we define the accumulated negative divergence $l_t$ as an \textbf{anchored feature-space expansion diagnostic}. This diagnostic monitors local volumetric expansion or contraction along imagined transitions. Large negative divergence indicates that downstream transitions are entering unconstrained or rapidly expanding regions of the learned latent feature space, serving as an empirical probe to halt return expansion before critic updates are corrupted by out-of-distribution dynamics.

\subsection{Objective and Scope of Boundary Selection}
\label{sec:objective_scope}

The algorithmic goal is to construct continuation gates $g_t \in [0, 1]$ from $l_t$ to modulate return expansion depth. A gate value of $g_t = 0$ truncates return expansion at step $t$ while retaining the critic's value estimate $V_\psi(x_t)$, shifting reliance from future model rollouts to the value function.

We explicitly distinguish conditional target variance at a fixed latent state from variance across replay samples or training seeds. Truncating return expansion avoids accumulating transition variance from downstream model rollouts at the cost of absorbing approximation bias $e(x_t) = V_\psi(x_t) - V^\pi(x_t)$ from the critic.

\subsection{Continuous-to-Discrete Bridging on Categorical Latents}
\label{sec:discrete_bridging}

Recurrent architectures like Dreamer V3 represent states using categorical latent variables concatenated with deterministic recurrent states. To bridge continuous transport concepts with discrete latent dynamics, we establish three theoretical properties:

\begin{enumerate}
    \item \textbf{Gaussian Smoothing over Categorical Feature Spaces:} Discrete distributions lack continuous density with respect to Lebesgue measure. Convolving the empirical posterior feature distribution $\mu_\phi$ with an isotropic Gaussian kernel $\mathcal{N}(0, \alpha^2 I_d)$ defines a smooth surrogate density:
    \begin{equation}
    p_{\phi,\alpha}(\tilde{x}) = \int_{\mathbb{R}^d} \mathcal{N}(\tilde{x}; x, \alpha^2 I_d) \, \mu_\phi(dx) \in C^\infty(\mathbb{R}^d).
    \end{equation}
    This smoothing guarantees that spatial score matching $s_\eta(\tilde{x}, \alpha) = \nabla_{\tilde{x}} \log p_{\phi,\alpha}(\tilde{x})$ and divergence operators $\nabla_x \cdot v_{\omega,\eta}(x)$ are smooth, well-defined differential operators over $\mathbb{R}^d$.

    \item \textbf{First-Order Quadrature over Latent Displacements:} An imagined transition $(x_t \to x_{t+1})$ represents a discrete step $\Delta u = 1$. Integrating the continuous rate $\frac{d}{du} \log \rho(x(u)) = -\nabla_x \cdot v(x(u))$ across a unit step yields $\log \rho(x_{t+1}) - \log \rho(x_t) = -\int_t^{t+1} \nabla_x \cdot v(x(u), u) \, du$. The discrete proxy update $\delta_t = \text{clip}(-\hat{D}_t, -C, C)$ implements a 1st-order forward Euler quadrature evaluated at $x_t$, with single-step discretization error bounded by $\mathcal{O}(\|\Delta x_t\|_2^2)$ where $\Delta x_t = x_{t+1} - x_t$.

    \item \textbf{Divergence as Local Volumetric Expansion Diagnostic:} A 1st-order spatial expansion along the step displacement $\Delta x_t$ gives $\Delta \log p(x_t) \approx \nabla_x \log p(x_t)^\top \Delta x_t$. The spatial divergence $\nabla_x \cdot v(x_t) = \text{Tr}(J_v(x_t))$ measures the trace of the field Jacobian, serving as an isotropic diagnostic of local feature space expansion or contraction.
\end{enumerate}

\section{Bridge Between Continuous Score Fields and Discrete RSSM}
An imagined trajectory in a latent world model involves two distinct mathematical objects:
\begin{itemize}
    \item \textbf{Discrete Transition Measure ($\mathbb{P}_{\text{RSSM}}$):} The discrete path measure induced by the recurrent stochastic state-space model (RSSM) transition kernel $z_{t+1} \sim p_\phi(z_{t+1} \mid z_t, a_t)$.
    \item \textbf{Auxiliary Diagnostic Field ($v_{\omega,\eta}$):} A continuous vector field $v_{\omega,\eta}(u, x, a) = f_\omega(u, x, a) + \frac{\sigma}{2}\epsilon_\eta(x, \sigma)$ constructed over Gaussian-smoothed posterior features $p_{\phi,\alpha}$.
\end{itemize}

Because imagined state sequences $(x_0, x_1, \dots, x_H)$ are sampled directly from $\mathbb{P}_{\text{RSSM}}$ rather than generated by integrating $\dot{x} = v_{\omega,\eta}(u, x, a)$, the auxiliary field functions strictly as an \textbf{evaluative diagnostic probe} along RSSM trajectories.

We formally decompose the discrepancy between exact density transport and the empirical diagnostic through three structural error components:
\begin{enumerate}
    \item \textbf{Kernel Smoothing Displacement ($E_{\text{smooth}}$):} The shift introduced by evaluating score estimators trained on Gaussian-perturbed replay features $p_{\phi,\alpha}(\tilde{x}) = \int \mathcal{N}(\tilde{x}; x, \alpha^2 I_d) \mu_\phi(dx)$ at clean imagined feature points $x_t$.
    \item \textbf{Euler Quadrature Residual ($E_{\text{euler}}$):} The first-order truncation error resulting from evaluating the continuous trace integral $\int_t^{t+1} \nabla_x \cdot v(x(u), u) \, du$ via a single forward Euler evaluation at $x_t$, bounded by $\mathcal{O}(\|x_{t+1} - x_t\|_2^2)$.
    \item \textbf{Structural Path Drift Gap ($\epsilon_{\text{drift}}$):} The unmodeled divergence between the discrete RSSM transition kernel and the continuous auxiliary reference field:
    \begin{equation}
        \epsilon_{\text{drift}} = D_{\text{KL}}(\mathbb{P}_{\text{RSSM}} \parallel \mathbb{P}_{\text{Flow}}).
    \end{equation}
\end{enumerate}

Because $\epsilon_{\text{drift}}$ is an unmodeled constant stemming from architectural separation, the proxy $l_t$ is treated strictly as a heuristic feature-space expansion diagnostic rather than a calibrated path density.

\section{Density-Guided Bootstrap Boundaries}
\label{sec:method}

\subsection{Overview}
\label{subsec:method_overview}

Our framework augments imagined actor--critic learning with a trajectory-dependent decision about where to bootstrap. The stochastic world model generates latent trajectories as in Equation~\eqref{eq:imagined_process}. Two auxiliary networks, a displacement predictor and a noise-conditioned score estimator, define a vector field evaluated along those trajectories. Accumulated negative divergence of this field provides a relative density proxy. Hard or soft gates constructed from this proxy modify the return recursion and determine the contribution of imagined states to the learning objectives.

The auxiliary networks do not replace the world-model transition distribution. The environment reward and continuation predictors also retain their original meanings. The proposed modification concerns return expansion and sample weighting, rather than a new reward function or an additional environment termination condition.

\subsection{Learning the Auxiliary Models}
\label{subsec:auxiliary_models}

Let $\bar x_i=\operatorname{sg}(F_\phi(z_i))$ be a replay posterior feature, where $\operatorname{sg}$ denotes stop-gradient. For a replay sequence with $L$ states, let $b_i$ be the action that produces the transition from state $i$ to state $i+1$, and let $m_i$ indicate that the pair does not cross an episode reset. We train a displacement network $f_\omega(u_i,\bar x_i,b_i)$ with normalized sequence-position input $u_i=i/(L-1)$:
\begin{equation}
\mathcal{L}_{\mathrm{disp}}
=
\frac{
\sum_i m_i
\left\|
f_\omega(u_i,\bar x_i,b_i)
-
(\bar x_{i+1}-\bar x_i)
\right\|_2^2
}{
d\,\max(1,\sum_i m_i)
}.
\label{eq:displacement_loss}
\end{equation}
The sums include all sequence pairs in the batch. Actions are aligned with the transition they generated; in the replay storage convention used by our implementation, this corresponds to the action stored with the succeeding observation.

The target is a one-step feature displacement. Although the implementation names this network a latent ODE component, training does not integrate an ODE or divide the displacement by a continuous-time increment. The input $u_i$ is a sequence-position conditioning variable, not an integration step size.

The score network uses the noise-prediction parameterization from Section~\ref{subsec:score_formulation}. With detached replay features, noise scale $\alpha$ drawn uniformly from $[\alpha_{\min},\alpha_{\max}]$, and $\epsilon\sim\mathcal{N}(0,I_d)$, its implemented loss is
\begin{equation}
\mathcal{L}_{\mathrm{score}}
=
\mathbb{E}
\left[
\frac{1}{d}
\left\|
\epsilon_\eta(\bar x+\alpha\epsilon,\alpha)-\epsilon
\right\|_2^2
\right].
\label{eq:implemented_score_loss}
\end{equation}
This is an unweighted denoising loss: neither rewards nor density gates weight its samples. The combined model-training objective is
\begin{equation}
\mathcal{L}_{\mathrm{model}}
=
\mathcal{L}_{\mathrm{WM}}
+
\beta_{\mathrm{disp}}\mathcal{L}_{\mathrm{disp}}
+
\beta_{\mathrm{score}}\mathcal{L}_{\mathrm{score}},
\label{eq:joint_model_objective}
\end{equation}
where $\mathcal{L}_{\mathrm{WM}}$ is the base world-model objective. Detaching the auxiliary inputs prevents these losses from directly updating the encoder or latent transition model through their feature targets. In the implementation, the losses share a model optimizer and global gradient clipping, so an auxiliary-enabled control remains distinct from an unmodified baseline.

\subsection{Constructing the Relative Density Proxy}
\label{subsec:proxy_construction}

At a fixed positive scale $\sigma$, define the score-corrected auxiliary field
\begin{equation}
v_{\omega,\eta}(u,x,a)
=
f_\omega(u,x,a)
-
\frac{\sigma^2}{2}s_\eta(x,\sigma)
=
f_\omega(u,x,a)
+
\frac{\sigma}{2}\epsilon_\eta(x,\sigma).
\label{eq:effective_field}
\end{equation}
The second equality uses Equation~\eqref{eq:noise_score_relation} and specifies the sign and scale of the implemented correction. The probability-flow construction motivates this form, but $\sigma$ is an auxiliary hyperparameter, not a calibrated diffusion coefficient of the stochastic latent model. We evaluate the predictor at the clean imagined feature with conditioning value $\sigma$. Choosing $\sigma$ within the training noise range does not remove the distribution shift between corrupted replay features and clean imagined features.

For each imagined state, we estimate spatial divergence with a stochastic trace estimator:
\begin{equation}
\widehat D_t
=
\xi_t^\top
\left[
\nabla_x v_{\omega,\eta}(u_t,x_t,a_t)
\right]
\xi_t,
\qquad
\mathbb{E}[\xi_t\xi_t^\top]=I_d.
\label{eq:divergence_estimator}
\end{equation}
For example, independent Rademacher entries satisfy the stated covariance condition. The action and position input are held fixed when differentiating with respect to $x_t$; thus this is not the total derivative of a policy-composed closed-loop field. The derivative includes the spatial dependence of the score network. Stopping gradients through the score input before this calculation would remove its contribution to the divergence.

We construct bounded increments and accumulate them from a common reference:
\begin{equation}
\delta_t
=
\operatorname{clip}(-\widehat D_t,-C,C),
\qquad
\ell_0=0,
\qquad
\ell_{t+1}=\ell_t+\delta_t,
\quad t=0,\ldots,H-1.
\label{eq:proxy_accumulation}
\end{equation}
The increment cap $C>0$ limits the influence of an individual divergence estimate. It also changes the signal and can conceal differences between fields when clipping is frequent. Conditional unbiasedness of the trace estimate therefore does not imply unbiasedness of the clipped increment or of the resulting proxy.

We compute the required spatial derivatives before detaching $\ell_t$ from actor--critic optimization. No absolute starting-state density is estimated, and the proxy is not exponentiated in the gate construction. It measures an anchored accumulation along the imagined path, rather than an absolute ranking of replay-state likelihoods. The field's mismatch with the actual latent transition process and its dependence on representation scale remain relevant limitations.

\subsection{Hard and Soft Gates}
\label{subsec:hard_soft_gates}

Let $\tau$ be a density threshold and $T>0$ a soft-gate temperature. The candidate gates are
\begin{align}
\widetilde g_t^{\mathrm{hard}}
&=
\mathbf{1}\{\ell_t\geq\tau\},
\label{eq:hard_gate}\\
\widetilde g_t^{\mathrm{soft}}
&=
\frac{1}{1+\exp(-(\ell_t-\tau)/T)}.
\label{eq:soft_gate}
\end{align}
Let $m\in\{1,\ldots,H\}$ be the minimum number of retained transition-loss entries and $N_{\mathrm{warm}}$ the warmup duration in the implementation's agent-step units. At training step $N$, the applied gate is
\begin{equation}
g_t=
\begin{cases}
1, & N<N_{\mathrm{warm}}\ \text{or}\ t<m,\\
\widetilde g_t, & \text{otherwise}.
\end{cases}
\label{eq:gate_overrides}
\end{equation}
This warmup delays activation; it is not a gradual interpolation of gate values. Setting $N_{\mathrm{warm}}=0$ activates selection from the start.

Define survival through state $t$ as
\begin{equation}
S_t=\prod_{j=0}^{t}g_j.
\label{eq:survival_weights}
\end{equation}
For hard gates, the first zero gate makes all subsequent survival weights zero, even if the proxy later rises above the threshold. For soft gates, survival decreases multiplicatively along the trajectory. Consequently, several moderate gate values can strongly attenuate later contributions even without a single sharp threshold crossing. We use these soft weights deterministically; the implementation does not sample a stopping time.

A useful diagnostic is the effective number of transition-loss entries,
\begin{equation}
H_{\mathrm{eff}}=\sum_{t=0}^{H-1}S_t.
\label{eq:effective_horizon}
\end{equation}
It equals the retained count in hard mode and a fractional survival mass in soft mode. It excludes discounting and $\lambda$-mixing, so it is not a complete measure of the influence of future rewards on a target. A fixed-boundary control uses $g_t=\mathbf{1}\{t<K\}$ for a prescribed $K\leq H$.

\subsection{Density-Gated Return Recursion}
\label{subsec:gated_returns}

We insert the gate at the succeeding state into the return recursion:
\begin{equation}
G_t^g
=
\widehat r_{t+1}
+
d_{t+1}
\left[
(1-\lambda g_{t+1})V_\psi(x_{t+1})
+
\lambda g_{t+1}G_{t+1}^g
\right],
\qquad
G_H^g=V_\psi(x_H).
\label{eq:gated_return}
\end{equation}
The gate therefore modulates the effective trace coefficient, $\lambda g_{t+1}$, rather than the environment continuation prediction.

When $g_{t+1}=0$, the target becomes
\begin{equation}
G_t^g
=
\widehat r_{t+1}+d_{t+1}V_\psi(x_{t+1}),
\label{eq:boundary_bootstrap}
\end{equation}
retaining both the incoming reward and the discounted boundary value. When all gates equal one, Equation~\eqref{eq:gated_return} recovers Equation~\eqref{eq:base_lambda_return}. The terminal gate $g_H$ does not alter the final backup because $G_H^g$ already equals $V_\psi(x_H)$.

In hard mode, the first failed gate determines the bootstrap boundary for targets rooted in the surviving prefix. Targets for later states may still be computed in the full tensor, but their actor and critic loss weights are zero. This separates stopping return expansion from stopping trajectory generation.

\subsection{Actor--Critic Objectives and Gradient Handling}
\label{subsec:weighted_objectives}

Let $W_t$ denote the detached discount weight inherited from the base agent. We apply the combined weight
\begin{equation}
\Omega_t=\operatorname{sg}(W_t S_t).
\label{eq:combined_weight}
\end{equation}
For a batch of $B$ imagined trajectories, a distributional critic with predictive distribution $p_\psi^V$ is trained using
\begin{equation}
\mathcal{L}_V
=
\frac{1}{BH}
\sum_{b=1}^{B}\sum_{t=0}^{H-1}
\Omega_{b,t}
\left[
-\log p_\psi^V\bigl(\operatorname{sg}(G_{b,t}^g)
\mid\operatorname{sg}(x_{b,t})\bigr)
+
\mathcal{R}_{b,t}^{\mathrm{slow}}
\right].
\label{eq:weighted_critic}
\end{equation}
Here $\mathcal{R}^{\mathrm{slow}}$ is the base agent's optional slow-critic regularization term. In our implementation it is the negative log probability assigned by the current critic to the detached scalar prediction of the slow critic.

Let $\mathcal{A}_{b,t}$ denote the actor surrogate used by the base gradient estimator. The actor objective is
\begin{equation}
\mathcal{L}_\pi
=
-\frac{1}{BH}
\sum_{b=1}^{B}\sum_{t=0}^{H-1}
\left[
\Omega_{b,t}\mathcal{A}_{b,t}
+
\beta_{\mathrm{ent}}S_{b,t}
\mathcal{H}\bigl(\pi_\theta(\cdot\mid x_{b,t})\bigr)
\right].
\label{eq:weighted_actor}
\end{equation}
The entropy term receives survival weighting; it does not receive the additional $W_t$ factor in the implemented base objective.

For continuous-control experiments using dynamics gradients, the surrogate is the return-minus-baseline difference divided by the base return-normalization scale, with differentiation through the retained imagination computation. For discrete-action experiments using a likelihood-ratio estimator, the implemented surrogate is
\begin{equation}
\mathcal{A}_{t}^{\mathrm{LR}}
=
\log\pi_\theta(a_t\mid\operatorname{sg}(x_t))\,
\operatorname{sg}\bigl(G_t^g-V_\psi(x_t)\bigr).
\label{eq:likelihood_ratio_surrogate}
\end{equation}
This likelihood-ratio expression concerns the policy distribution and is distinct from the latent-density score estimator.

Gates and survival weights are detached for both gradient estimators. Thus actor updates do not differentiate through the threshold, sigmoid, or density diagnostic. The critic targets and critic input features are also detached in the critic update. During dynamics-based actor learning, world-model and auxiliary parameters remain fixed while the base differentiable imagination paths are retained.

The denominator in Equations~\eqref{eq:weighted_critic} and \eqref{eq:weighted_actor} is the full batch-time size $BH$, not the surviving weight mass. Shorter effective horizons can therefore change update magnitude as well as sample selection. The implementation also excludes zero-weight targets from return-scale statistics in hard and fixed modes, whereas soft mode retains all targets for those statistics. These choices must be reported and controlled when attributing improvements to density-dependent selection.

\subsection{Algorithm and DreamerV3 Instantiation}
\label{subsec:method_algorithm}

Algorithm~\ref{alg:density_guided_learning} summarizes the
training procedure. The counter $N$ uses the implementation's
agent-step convention. The stochastic world model generates
the full imagined trajectory before boundary selection.
The spatial derivative includes the score network's
dependence on its input; detachment is applied only after
the divergence-based proxy has been computed.

\begin{algorithm}[!hbt]
\caption{Density-Guided Imagined Actor-Critic Learning}
\label{alg:density_guided_learning}
\begin{algorithmic}[1]
\REQUIRE World model parameters $\phi$; actor $\pi_\theta$; critic $V_\psi$; displacement field $f_\omega$; noise predictor $\epsilon_\eta$; replay buffer $\mathcal{D}$.
\REQUIRE Imagination horizon $H$; trace coefficient $\lambda$; threshold $\tau$; temperature $T$; increment cap $C$; score scale $\sigma$; minimum retained depth $m$; activation step $N_{\text{warm}}$; boundary mode $\mathcal{M} \in \{\text{hard}, \text{soft}\}$.
\WHILE{interaction budget remains}
    \STATE Collect environment experience using $\pi_\theta$ and append it to $\mathcal{D}$.
    \STATE Sample replay sequences and infer posterior latent features.
    \STATE Update world model and auxiliary parameters using $\mathcal{L}_{\text{WM}} + \beta_{\text{disp}}\mathcal{L}_{\text{disp}} + \beta_{\text{score}}\mathcal{L}_{\text{score}}$ (Algorithm~\ref{alg:wm_update}).
    \STATE Start from detached replay posterior states and generate full stochastic imagined trajectory $x_0, a_0, x_1, \dots, a_{H-1}, x_H$.
    \STATE Predict rewards $\hat{r}_{t+1}$, discounts $d_{t+1} = \gamma \hat{c}_{t+1}$, and values $V_\psi(x_t)$.
    \STATE Set $l_0 \gets 0$.
    \FOR{$t = 0, \dots, H-1$}
        \STATE Define $v_t(x) \gets f_\omega(u_t, x, a_t) + \frac{\sigma}{2} \epsilon_\eta(x, \sigma)$ with $u_t = t / H$.
        \STATE Draw an independent trace probe $\xi_t$ satisfying $\mathbb{E}[\xi_t \xi_t^\top] = \mathbf{I}$.
        \STATE Estimate $\hat{D}_t \gets \xi_t^\top [\nabla_x v_t(x)]_{x=x_t} \xi_t$, holding $a_t$ and $u_t$ fixed.
        \STATE Set $l_{t+1} \gets l_t + \text{clip}(-\hat{D}_t, -C, C)$.
    \ENDFOR
    \STATE Detach density proxies $l_{0:H}$ from actor-critic optimization.
    \FOR{$t = 0, \dots, H$}
        \IF{$N < N_{\text{warm}}$ \OR $t < m$}
            \STATE $g_t \gets 1$.
        \ELSIF{$\mathcal{M} = \text{hard}$}
            \STATE $g_t \gets \mathbb{I}(l_t \ge \tau)$.
        \ELSE
            \STATE $g_t \gets [1 + \exp(-(l_t - \tau) / T)]^{-1}$.
        \ENDIF
    \ENDFOR
    \STATE Set $S_0 \gets g_0$.
    \FOR{$t = 1, \dots, H$}
        \STATE $S_t \gets S_{t-1} g_t$.
    \ENDFOR
    \STATE Set $G_H \gets V_\psi(x_H)$.
    \FOR{$t = H-1, \dots, 0$}
        \STATE $G_t \gets \hat{r}_{t+1} + d_{t+1} [(1 - \lambda g_{t+1}) V_\psi(x_{t+1}) + \lambda g_{t+1} G_{t+1}]$.
    \ENDFOR
    \STATE Form detached loss weights $\Omega_t \gets W_t S_t$ for $t = 0, \dots, H-1$ using base agent discount weights $W_t$.
    \STATE Update actor and critic using gated returns and survival-weighted objectives (Section 5.6).
    \STATE Apply base agent's slow-critic update and advance training counters.
\ENDWHILE
\end{algorithmic}
\end{algorithm}


The general formulation requires a latent feature representation, imagined transitions, and a bootstrap value estimator. Our implementation uses DreamerV3's recurrent stochastic state-space model and distributional critic \citep{hafner2023dreamerv3}. The auxiliary networks operate on the concatenation of deterministic and stochastic latent features. In the configuration with a 512-dimensional deterministic component and 32 categorical variables with 32 categories each, this feature vector has dimension $512+32\cdot32=1536$.

Section~\ref{sec:problem_formulation} defines $H$ as a transition count. The implementation's \texttt{imag\_horizon} counts feature entries: setting it to 15 produces 14 return-training transitions, corresponding to $H=14$ here. Its auxiliary position input during imagination is $u_t=t/H$, while replay uses $i/(L-1)$. These are sequence-relative conditioning choices, not a shared physical-time discretization.

For implementation fidelity, let $\widehat c_t$ denote the continuation prediction at feature entry $t$, including the initial entry. The inherited discount-loss weights are
\begin{equation}
W_0=1,
\qquad
W_t=\prod_{j=0}^{t-1}(\gamma\widehat c_j),
\quad t\geq1.
\label{eq:implementation_discount_weights}
\end{equation}
The return recursion separately uses the next-state predictions $d_{t+1}=\gamma\widehat c_{t+1}$. We preserve this indexing rather than silently changing the base agent's discount weights. Warmup is measured by the agent-step counter. With action repeat two, 100,000 agent steps correspond to 200,000 logged environment steps. Auxiliary training can proceed during this period while all gates remain one. Minimum-attention regularization, direct density reward weighting, and learned auxiliary diffusion are disabled in the boundary experiments described here. The chosen threshold, temperature, increment cap, and activation time are experimental settings, not universal constants of the method. Finally, all imagined states are generated before gating. The method adapts the portion of imagination used in return expansion and optimization; this implementation does not claim computational savings from early rollout termination. 
Appendix~\ref{app:dreamer_implementation} provides
DreamerV3-specific pseudocode, including auxiliary training,
RSSM imagination, gradient handling, and loss weighting.

\section{Theoretical Analysis}
\label{sec:theoretical_analysis}


\subsection*{6.1 Augmented Markov State Representation}

In our gated return recursion, the continuation gate $g_{t+1} = g(l_{t+1})$ depends on the accumulated divergence proxy $l_{t+1} = l_t + \delta_t$, where $\delta_t = \mathrm{clip}(-\hat{D}_t, -C, C)$ is computed from $(u_t, x_t, a_t)$ and an independent Rademacher probe $\xi_t$. Because $l_{t+2} = l_{t+1} + \delta_{t+1}$ propagates the proxy history forward, $G_{t+1}^g$ is not conditionally independent of $g_{t+1}$ given $x_{t+1}$ alone.

To obtain a Markovian representation of the gated return process, we define the \emph{augmented state} $\bar{x}_t \triangleq (x_t, l_t) \in \mathcal{S} \triangleq \mathcal{X} \times \mathbb{R}$, initialized at $l_0 = 0$. Its transition kernel $p_{\phi, v}(\bar{x}_{t+1} \mid \bar{x}_t, a_t)$ is induced by $z_{t+1} \sim p_\phi(\cdot \mid z_t, a_t)$, $x_{t+1} = F_\phi(z_{t+1})$, and $l_{t+1} = l_t + \mathrm{clip}(-\hat{D}_t, -C, C)$ after marginalizing over the independent trace probe $\xi_t$. We also denote the one-step transition filtration by $\mathcal{F}_{t+1} \triangleq \sigma(\bar{x}_t, a_t, \xi_t, x_{t+1})$, under which $\hat{r}_{t+1}$, $d_{t+1}$, $x_{t+1}$, and $g_{t+1} = g(l_{t+1})$ are $\mathcal{F}_{t+1}$-measurable, while $G_{t+1}^g$ depends on $\mathcal{F}_{t+1}$ solely through $\bar{x}_{t+1} = (x_{t+1}, l_{t+1})$.

\begin{rem}[Diagnostic Nature of the Score Proxy]\label{rem:proxy_diagnostic}
The accumulated signal $l_t = \sum_{j=0}^{t-1} \delta_j$ tracks local volumetric expansion and contraction of the auxiliary vector field $v_{\omega,\eta}$ evaluated at states sampled from $\mathbb{P}_{\mathrm{RSSM}}$. It does not quantify exact path probabilities or absolute likelihoods under the world model's discrete stochastic transition kernel.
\end{rem}

\begin{rem}[Role of the Path Drift Gap]\label{rem:drift_gap}
The structural path drift gap $\epsilon_{\mathrm{drift}} = D_{\mathrm{KL}}(\mathbb{P}_{\mathrm{RSSM}} \,\|\, \mathbb{P}_{\mathrm{Flow}})$ represents an irreducible discrepancy between the stochastic world-model rollout measure and the continuous reference flow. The utility of $l_t$ for gating return expansion relies on detecting local feature-space divergence along imagined trajectories rather than eliminating $\epsilon_{\mathrm{drift}}$.
\end{rem}

\begin{prop}[Denoising Score Identity for Gaussian-Smoothed Features]\label{prop:score}
Let $\mu_\phi$ be the empirical probability measure of posterior replay features on $\mathbb{R}^d$. For any fixed noise scale $\sigma > 0$, define the Gaussian-smoothed marginal density
\begin{equation}
p_{\phi,\sigma}(\tilde{x}) = \int_{\mathbb{R}^d} \mathcal{N}(\tilde{x}; x, \sigma^2 I_d)\,\mu_\phi(dx) \in C^\infty(\mathbb{R}^d),
\end{equation}
which is strictly positive and infinitely differentiable everywhere on $\mathbb{R}^d$. If $\epsilon_\eta^*(\cdot, \sigma)$ minimizes the population denoising objective $\mathcal{L}_{\mathrm{score}}$ at scale $\sigma$, then for every $\tilde{x} \in \mathbb{R}^d$,
\begin{equation}\label{eq:score_param}
s_\eta^*(\tilde{x}, \sigma) \triangleq -\frac{\epsilon_\eta^*(\tilde{x}, \sigma)}{\sigma} = \nabla_{\tilde{x}} \log p_{\phi,\sigma}(\tilde{x}).
\end{equation}
Moreover, in the special case where $\mu_\phi$ admits a density $p \in C^1(\mathcal{U})$ that is strictly positive on an open set $\mathcal{U} \subseteq \mathbb{R}^d$, $\lim_{\sigma \to 0^+} s_\eta^*(x, \sigma) = \nabla_x \log p(x)$ for all $x \in \mathcal{U}$; for discrete or finite-sample replay distributions, $\sigma > 0$ serves strictly as a smooth mollifier.
\end{prop}

\begin{proof}
For $\tilde{x} = x + \sigma\epsilon$ with $x \sim \mu_\phi$ and $\epsilon \sim \mathcal{N}(0, I_d)$, minimizing $\mathbb{E}_{x,\epsilon}[\|\epsilon_\eta(x + \sigma\epsilon, \sigma) - \epsilon\|_2^2]$ yields the conditional expectation $\epsilon_\eta^*(\tilde{x}, \sigma) = \mathbb{E}[\epsilon \mid \tilde{x}] = \mathbb{E}[(\tilde{x} - x)/\sigma \mid \tilde{x}]$. Because the Gaussian kernel $\mathcal{N}(\tilde{x}; x, \sigma^2 I_d)$ is smooth and rapidly decaying, differentiating under the integral sign gives
\begin{equation}
\nabla_{\tilde{x}} p_{\phi,\sigma}(\tilde{x}) = \int_{\mathbb{R}^d} -\frac{\tilde{x} - x}{\sigma^2}\,\mathcal{N}(\tilde{x}; x, \sigma^2 I_d)\,\mu_\phi(dx) = -\frac{p_{\phi,\sigma}(\tilde{x})}{\sigma}\,\mathbb{E}\!\left[\frac{\tilde{x} - x}{\sigma} \;\middle|\; \tilde{x}\right] = -\frac{p_{\phi,\sigma}(\tilde{x})}{\sigma}\,\epsilon_\eta^*(\tilde{x}, \sigma).
\end{equation}
Dividing by $p_{\phi,\sigma}(\tilde{x}) > 0$ establishes~\eqref{eq:score_param}. When $\mu_\phi$ admits a $C^1$ density $p > 0$ on $\mathcal{U}$, standard approximate-identity convergence gives $p_{\phi,\sigma} \to p$ and $\nabla p_{\phi,\sigma} \to \nabla p$ locally uniformly on $\mathcal{U}$ as $\sigma \to 0^+$, yielding $\lim_{\sigma \to 0^+} s_\eta^*(x, \sigma) = \nabla_x \log p(x)$.
\end{proof}

\begin{theo}[Recursive Critic Bias Propagation]\label{thm:bias}
Let $\bar{x}_t = (x_t, l_t)$ be the augmented state at imagination step $t$, let $e(x) \triangleq V_\psi(x) - V^\pi(x)$ denote the critic approximation bias, and define the conditional target bias $\mathrm{Bias}(G_t^g \mid \bar{x}_t) \triangleq \mathbb{E}[G_t^g \mid \bar{x}_t] - V^\pi(x_t)$. Assuming $V^\pi(x_t) = \mathbb{E}[\hat{r}_{t+1} + d_{t+1}V^\pi(x_{t+1}) \mid x_t]$ under the world model, the conditional bias satisfies the recursion
\begin{equation}\label{eq:bias_recursion}
\mathrm{Bias}(G_t^g \mid \bar{x}_t) = \mathbb{E}\!\big[d_{t+1}(1 - \lambda g_{t+1}) e(x_{t+1}) \;\big|\; \bar{x}_t\big] + \mathbb{E}\!\big[d_{t+1}\lambda g_{t+1} \mathrm{Bias}(G_{t+1}^g \mid \bar{x}_{t+1}) \;\big|\; \bar{x}_t\big].
\end{equation}
\end{theo}

The proof of Theorem~\ref{thm:bias} is given in Appendix~A.

\begin{theo}[Auxiliary Divergence Estimation Error under Path Drift]\label{thm:trace}
Let $J_v(u_t, x_t, a_t) \triangleq \nabla_x v_{\omega,\eta}(u_t, x_t, a_t) \in \mathbb{R}^{d \times d}$ be the spatial Jacobian of the auxiliary score-corrected vector field. Assume there exist constants $M, F > 0$ such that $|\mathrm{Tr}(J_v(u_t, x_t, a_t))| \le M$ and $\|J_v(u_t, x_t, a_t)\|_F \le F$ uniformly over all reachable $(x_t, a_t)$. Let $\mathbb{P}_{\mathrm{RSSM}, t}$ and $\mathbb{P}_{\mathrm{Flow}, t}$ denote the step-$t$ marginal distributions of $(x_t, a_t)$ under the stochastic world model and reference flow, respectively, and assume the trajectory-level KL divergence satisfies $D_{\mathrm{KL}}(\mathbb{P}_{\mathrm{RSSM}} \,\|\, \mathbb{P}_{\mathrm{Flow}}) \le \epsilon_{\mathrm{drift}}$. Then, for an independent Rademacher probe $\xi_t \in \{-1, +1\}^d$ in $\hat{D}_t = \xi_t^\top J_v(u_t, x_t, a_t)\xi_t$, with probability at least $1 - \delta$ (jointly over $(x_t, a_t) \sim \mathbb{P}_{\mathrm{RSSM}, t}$ and $\xi_t$),
\begin{align}\label{eq:trace_bound}
\Delta_t &\triangleq \Big|\hat{D}_t - \mathbb{E}_{\mathbb{P}_{\mathrm{Flow}, t}}[\mathrm{Tr}(J_v(u_t, x_t, a_t))]\Big| \nonumber \\
&\le \sqrt{\frac{2\mathbb{E}_{\mathbb{P}_{\mathrm{RSSM}, t}}[\|J_v\|_F^2] + \mathrm{Var}_{\mathbb{P}_{\mathrm{RSSM}, t}}(\mathrm{Tr}(J_v))}{\delta}} + M\sqrt{2\epsilon_{\mathrm{drift}}} \le \sqrt{\frac{2F^2 + M^2}{\delta}} + M\sqrt{2\epsilon_{\mathrm{drift}}}.
\end{align}
\end{theo}

\begin{proof}
We decompose the total error $\Delta_t$ via the triangle inequality into a stochastic estimation error under $\mathbb{P}_{\mathrm{RSSM}, t}$ and a deterministic distributional drift error:
\begin{equation}\label{eq:trace_decomp}
\Delta_t \le \Big|\hat{D}_t - \mathbb{E}_{\mathbb{P}_{\mathrm{RSSM}, t}}[\mathrm{Tr}(J_v)]\Big| + \Big|\mathbb{E}_{\mathbb{P}_{\mathrm{RSSM}, t}}[\mathrm{Tr}(J_v)] - \mathbb{E}_{\mathbb{P}_{\mathrm{Flow}, t}}[\mathrm{Tr}(J_v)]\Big|.
\end{equation}
For the first term, conditioned on $(x_t, a_t)$, the Hutchinson Rademacher trace estimator satisfies $\mathbb{E}_{\xi_t}[\hat{D}_t \mid x_t, a_t] = \mathrm{Tr}(J_v)$ and $\mathrm{Var}_{\xi_t}(\hat{D}_t \mid x_t, a_t) = 2\big(\|J_v\|_F^2 - \sum_{i=1}^d [J_v]_{ii}^2\big) \le 2\|J_v\|_F^2$. Taking expectations over $(x_t, a_t) \sim \mathbb{P}_{\mathrm{RSSM}, t}$ gives $\mathbb{E}[\hat{D}_t] = \mathbb{E}_{\mathbb{P}_{\mathrm{RSSM}, t}}[\mathrm{Tr}(J_v)]$, and the Law of Total Variance yields
\begin{align}
\mathrm{Var}(\hat{D}_t) &= \mathbb{E}_{\mathbb{P}_{\mathrm{RSSM}, t}}\!\big[\mathrm{Var}_{\xi_t}(\hat{D}_t \mid x_t, a_t)\big] + \mathrm{Var}_{\mathbb{P}_{\mathrm{RSSM}, t}}\!\big(\mathrm{Tr}(J_v)\big) \nonumber \\
&\le 2\mathbb{E}_{\mathbb{P}_{\mathrm{RSSM}, t}}[\|J_v\|_F^2] + \mathrm{Var}_{\mathbb{P}_{\mathrm{RSSM}, t}}(\mathrm{Tr}(J_v)) \le 2F^2 + M^2,
\end{align}
since $\mathrm{Var}_{\mathbb{P}_{\mathrm{RSSM}, t}}(\mathrm{Tr}(J_v)) \le \mathbb{E}_{\mathbb{P}_{\mathrm{RSSM}, t}}[\mathrm{Tr}(J_v)^2] \le M^2$. Applying Chebyshev's inequality, for any $\epsilon > 0$,
\begin{equation}
\mathbb{P}\!\Big(\big|\hat{D}_t - \mathbb{E}_{\mathbb{P}_{\mathrm{RSSM}, t}}[\mathrm{Tr}(J_v)]\big| \ge \epsilon\Big) \le \frac{\mathrm{Var}(\hat{D}_t)}{\epsilon^2} \le \frac{2\mathbb{E}_{\mathbb{P}_{\mathrm{RSSM}, t}}[\|J_v\|_F^2] + \mathrm{Var}_{\mathbb{P}_{\mathrm{RSSM}, t}}(\mathrm{Tr}(J_v))}{\epsilon^2}.
\end{equation}
Setting the right-hand side equal to $\delta$ and solving for $\epsilon$ shows that, with probability at least $1 - \delta$,
\begin{equation}\label{eq:trace_est_part}
\Big|\hat{D}_t - \mathbb{E}_{\mathbb{P}_{\mathrm{RSSM}, t}}[\mathrm{Tr}(J_v)]\Big| \le \sqrt{\frac{2\mathbb{E}_{\mathbb{P}_{\mathrm{RSSM}, t}}[\|J_v\|_F^2] + \mathrm{Var}_{\mathbb{P}_{\mathrm{RSSM}, t}}(\mathrm{Tr}(J_v))}{\delta}} \le \sqrt{\frac{2F^2 + M^2}{\delta}}.
\end{equation}

For the second term in~\eqref{eq:trace_decomp}, the Data Processing Inequality for KL divergence implies $D_{\mathrm{KL}}(\mathbb{P}_{\mathrm{RSSM}, t} \,\|\, \mathbb{P}_{\mathrm{Flow}, t}) \le D_{\mathrm{KL}}(\mathbb{P}_{\mathrm{RSSM}} \,\|\, \mathbb{P}_{\mathrm{Flow}}) \le \epsilon_{\mathrm{drift}}$. By Pinsker's inequality, the total variation distance satisfies
\begin{equation}
\delta_{\mathrm{TV}}(\mathbb{P}_{\mathrm{RSSM}, t}, \mathbb{P}_{\mathrm{Flow}, t}) \le \sqrt{\frac{1}{2}D_{\mathrm{KL}}(\mathbb{P}_{\mathrm{RSSM}, t} \,\|\, \mathbb{P}_{\mathrm{Flow}, t})} \le \sqrt{\frac{\epsilon_{\mathrm{drift}}}{2}}.
\end{equation}
Because $|\mathrm{Tr}(J_v)| \le M$ everywhere, the difference in expectations is bounded by $2M\,\delta_{\mathrm{TV}}(\mathbb{P}_{\mathrm{RSSM}, t}, \mathbb{P}_{\mathrm{Flow}, t})$:
\begin{equation}\label{eq:trace_drift_part}
\Big|\mathbb{E}_{\mathbb{P}_{\mathrm{RSSM}, t}}[\mathrm{Tr}(J_v)] - \mathbb{E}_{\mathbb{P}_{\mathrm{Flow}, t}}[\mathrm{Tr}(J_v)]\Big| \le 2M\,\delta_{\mathrm{TV}}(\mathbb{P}_{\mathrm{RSSM}, t}, \mathbb{P}_{\mathrm{Flow}, t}) \le M\sqrt{2\epsilon_{\mathrm{drift}}}.
\end{equation}
Substituting~\eqref{eq:trace_est_part} and~\eqref{eq:trace_drift_part} into~\eqref{eq:trace_decomp} completes the proof.
\end{proof}

\section{Experimental Setup}
\label{sec:experimental_setup}

Our experiments investigate whether density-guided bootstrap selection improves learning beyond the effects of auxiliary training and shorter imagined returns. All variants use the same modified DreamerV3 implementation. We separate the choice of boundary rule from the world-model architecture, interaction budget, and optimization settings.


\subsection{Environments and Observation Settings}
\label{subsec:experimental_environments}

The continuous-control study considers the DeepMind Control Suite tasks \texttt{cheetah\_run}, \texttt{walker\_walk}, \texttt{walker\_run}, and \texttt{finger\_turn\_hard}. These experiments use vector observations through the \texttt{dmc\_proprio} configuration, with multilayer-perceptron encoders and decoders. Cheetah refers specifically to the Control Suite task, rather than the Gym MuJoCo HalfCheetah benchmark. The discrete-action study uses Atari Pong with $64\times64$ RGB observations and the \texttt{atari100k} configuration. This setting provides a complementary test involving visual representation learning and a categorical action policy. Table~\ref{tab:environment_protocol} summarizes the interaction settings.

\begin{table}[t]
\centering
\small
\begin{tabular}{llll}
\hline
Setting & Observations & Action repeat & Interaction budget \\
\hline
Control Suite & Proprioceptive vectors & 2 & 500,000 environment steps \\
Atari Pong & $64\times64$ RGB images & 4 & 400,000 emulator frames \\
\hline
\end{tabular}
\caption{Observation and interaction settings. Budgets follow the repeat-adjusted logging convention, corresponding nominally to 250,000 agent decisions for Control Suite and 100,000 for Pong.}
\label{tab:environment_protocol}
\end{table}

The main comparison evaluates vanilla DreamerV3 and the
density-guided hard and soft variants using 10 independent
training seeds per method and task. Auxiliary and
fixed-boundary controls are reported separately in
Sections \ref{subsec:fixed_boundary_results} and \ref{subsec:component_ablation_results}. The boundary-behavior illustrations
use one supplied training run per variant, and the threshold-sensitivity study uses 10 seeds. The Cheetah Run threshold sweep and the additional controls for score correction, boundary assignment, and loss weighting uses 10 seeds.


\subsection{Reference Agents and Controlled Comparisons}
\label{subsec:experimental_baselines}

The evaluation design includes the following variants:
\begin{enumerate}
\item \textbf{Vanilla DreamerV3.}
The base return construction is retained. Auxiliary displacement and score training, density selection, direct density weighting, and minimum-attention regularization are disabled.
\item \textbf{Auxiliary-enabled, cutoff off.}
The auxiliary networks are trained and the density diagnostic is computed, but no density boundary modifies returns or loss weights. This control is distinct from vanilla DreamerV3.
\item \textbf{Density-guided hard boundary.}
The score-corrected proxy selects a hard bootstrap boundary, with the survival-weighted objectives defined in Section~\ref{sec:method}.
\item \textbf{Density-guided soft boundary.}
The same proxy supplies sigmoid continuation gates and cumulative survival weights. This changes return expansion and loss weighting rather than directly multiplying rewards by density.
\item \textbf{Fixed bootstrap boundary.}
A prescribed boundary replaces trajectory-dependent selection while retaining the boundary value. To isolate the selection rule, auxiliary computation should match the density-guided variant. A separate fixed-boundary agent without auxiliaries measures the performance of the simpler alternative.
\end{enumerate}

These variants define the ablation design. 
Matching a fixed boundary to the mean effective horizon does not match its distribution or evolution during training. If the adaptive horizon changes substantially, a predetermined fixed-horizon schedule provides an additional control. Any schedule derived from a reference run must be frozen before evaluation and disclosed.

\subsection{Training Configuration}
\label{subsec:experimental_training}

Within each comparison, we retain the same architecture, replay settings, world-model losses, optimizers, and interaction budget unless an explicitly named ablation changes them. The supplied configuration uses batch size 16, replay sequence length 64, discount 0.997, and return-mixing coefficient 0.95. An imagination setting of 15 feature entries produces 14 return-training transitions.

The auxiliary setup uses $\sigma=0.05$, denoising scales sampled from $[0.01,0.10]$, displacement-loss coefficient $10^{-3}$, and score-loss coefficient $10^{-4}$. Learned auxiliary diffusion, direct density reward weighting, Liouville reward weighting, and minimum-attention regularization are disabled.

The Control Suite configuration uses four environments, action repeat two, dynamics-based actor gradients, and training-ratio parameter 512. The Atari configuration uses one environment, action repeat four, likelihood-ratio actor gradients, and training-ratio parameter 1024. Training-ratio values are implementation configuration parameters whose meaning depends on the update scheduler. Compilation is disabled in the experimental commands.


\subsection{Boundary Parameters and Sensitivity Studies}
\label{subsec:experimental_boundary_settings}

The current density-boundary experiments retain a minimum of $m=2$ transition-loss entries. The Cheetah threshold study varies $\tau$ over $\{-500,-700,-900\}$ for hard and soft modes. It holds the soft temperature at $T=100$, increment cap at $C=200$, each figure caption states which setting was used.

This threshold sweep is exploratory and uses 10 seeds. 

Proxy scales can differ across tasks and training stages. Diagnostic sweeps report proxy quantiles across imagination depths and the induced effective-horizon distribution. When diagnostics determine parameters, report the reference checkpoint, replay sample, candidate set, and selection rule. Selection based on an auxiliary-enabled reference run should be distinguished from tuning against final evaluation returns.

Immediate-activation and delayed-activation experiments are reported separately. Warmup delays selection rather than gradually ramping it. For Atari with action repeat four, activation at 200,000 frames would require 50,000 agent steps; the Control Suite counter value would imply a different activation point.


\subsection{Evaluation and Statistical Reporting}
\label{subsec:experimental_evaluation}

The primary measure is undiscounted evaluation episode return versus environment interaction. Evaluation episodes do not update the model or policy. All variants in a comparison use the same evaluation-policy convention, episode count, evaluation frequency, and time-limit handling. These settings are recovered from resolved run configurations rather than inferred from defaults. We report performance at a common interaction budget. A predefined average over the final evaluation checkpoints may additionally summarize late-training performance, but it is labeled as a within-run temporal average. It is not an independent-seed estimate. If collection or logging extends beyond the nominal budget, budget-matched summaries consistently exclude those later observations.
We evaluate each method included in the main performance comparison
using 10 independent training seeds. At each evaluation checkpoint,
episode returns are first averaged within each training run. We then
report the mean evaluation return across the 10 runs. Training seeds,
rather than evaluation episodes or successive checkpoints, are the
independent units used to quantify run-to-run variability. The learning curves summarize the mean evaluation return
across 10 independent training runs. Shaded regions are
intended to represent one standard deviation across runs,
rather than confidence intervals for the mean. 

\subsection{Mechanism Diagnostics}
\label{subsec:experimental_diagnostics}

We inspect effective imagination length during training
to characterize the behavior of the boundary mechanism.
For hard boundaries, this metric counts retained
transition-loss entries. For soft boundaries, it sums
survival weights. The illustrations in Section \ref{subsec:boundary_behavior} use
one training run per variant and task and do not estimate
variation across training seeds. Additional logged diagnostics include proxy quantiles,
increment-clipping fractions, and trial cutoff-length
statistics. Trial length standard deviations describe
variation across imagined trajectories within diagnostic
batches; they are distinct from evaluation-return
variability across training seeds. The implementation generates the full imagined trajectory
before applying the gates. Consequently, reduced effective
imagination length is not interpreted as reduced
trajectory-generation cost. We use three additional controls on Cheetah Run to distinguish how the boundary is computed, which trajectory receives it, and how much weight the retained transitions contribute to the loss. 
All three comparisons use a hard boundary and activate selection at 200,000 logged
environment steps; the thresholds are $\tau=-900$ for the score-correction control,
$\tau=-700$ for the boundary-assignment control, and $\tau=-700$ for the
loss-normalization control. In the \emph{drift-only} control, both auxiliary networks are trained, but the score correction is omitted when computing the divergence used by the gate. In the \emph{boundary-assignment} control, two runs start
from the same 200,000-step checkpoint. The matched variant applies each trajectory's own boundary; the shuffled variant permutes the boundary sequences across trajectories within each batch. This preserves the set of boundaries in that batch while changing which trajectory receives each one. In the \emph{mass-compensation} control, the score-corrected boundary rule is unchanged, but the surviving loss weights are rescaled to compensate for the reduction in total weight caused by gating.

\section{Results and Analysis}
\label{sec:results}

We organize the evaluation around four questions: whether density-guided boundaries improve performance relative to vanilla DreamerV3, whether the gains extend beyond a fixed bootstrap boundary, how enabling the auxiliary mechanism differs from using its signal for gating, and how performance depends on the density threshold. The main learning curves summarize 10 independent training seeds. Shaded regions show one standard deviation across seeds, and TensorBoard-style exponential smoothing with weight 0.7 is used for visualization. These bands describe run-to-run dispersion, not confidence intervals for the mean.


\subsection{Performance Across Environments}
\label{subsec:main_results}

\begin{figure}[t]
\centering
  \includegraphics[width=0.32\linewidth]{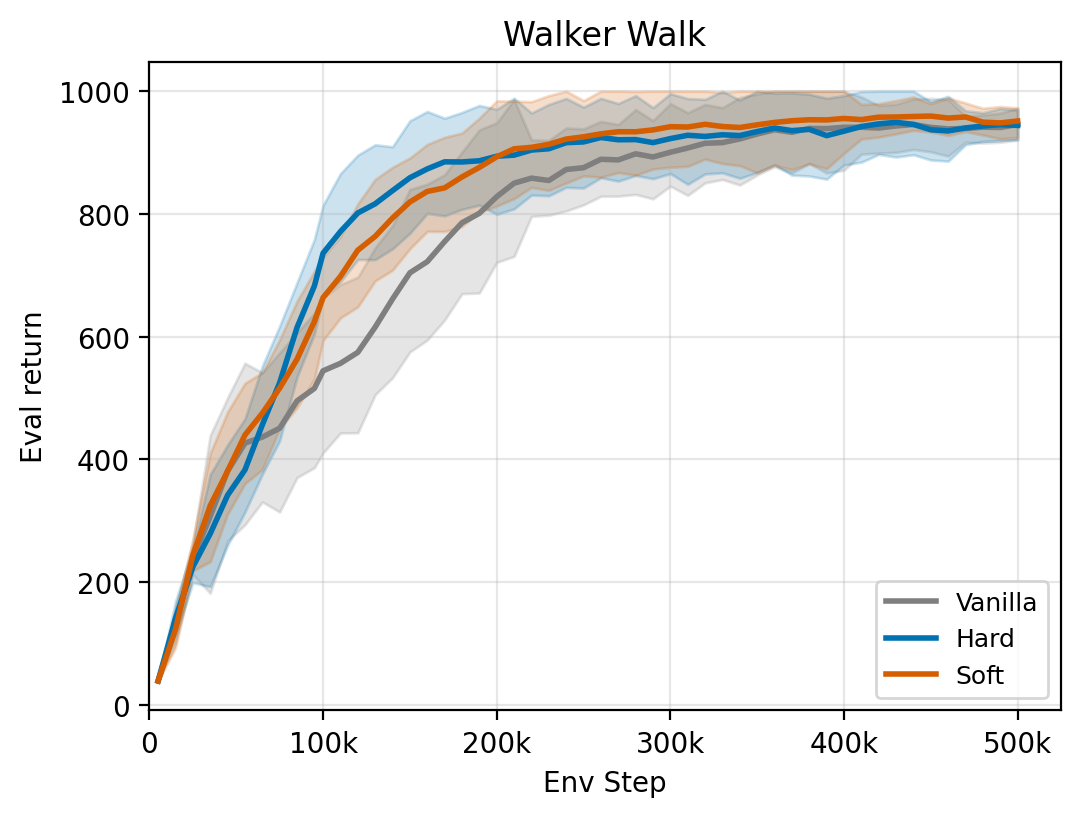}
  \includegraphics[width=0.32\linewidth]{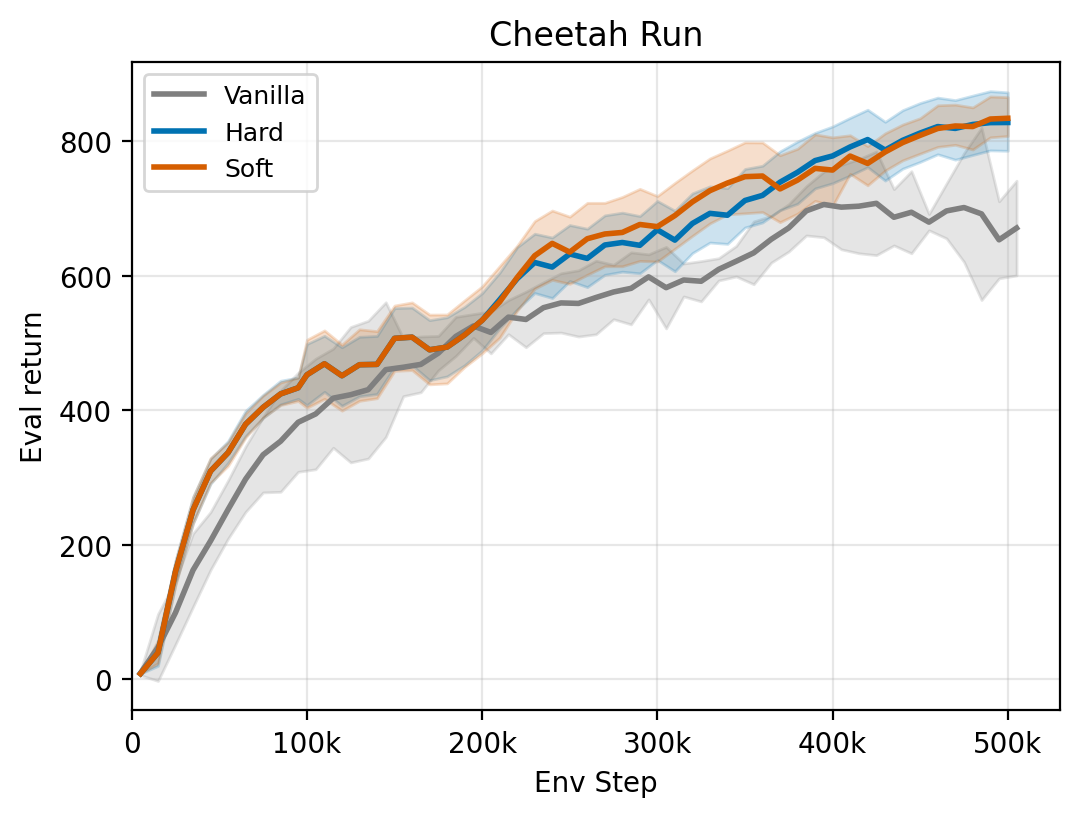}
  \includegraphics[width=0.32\linewidth]{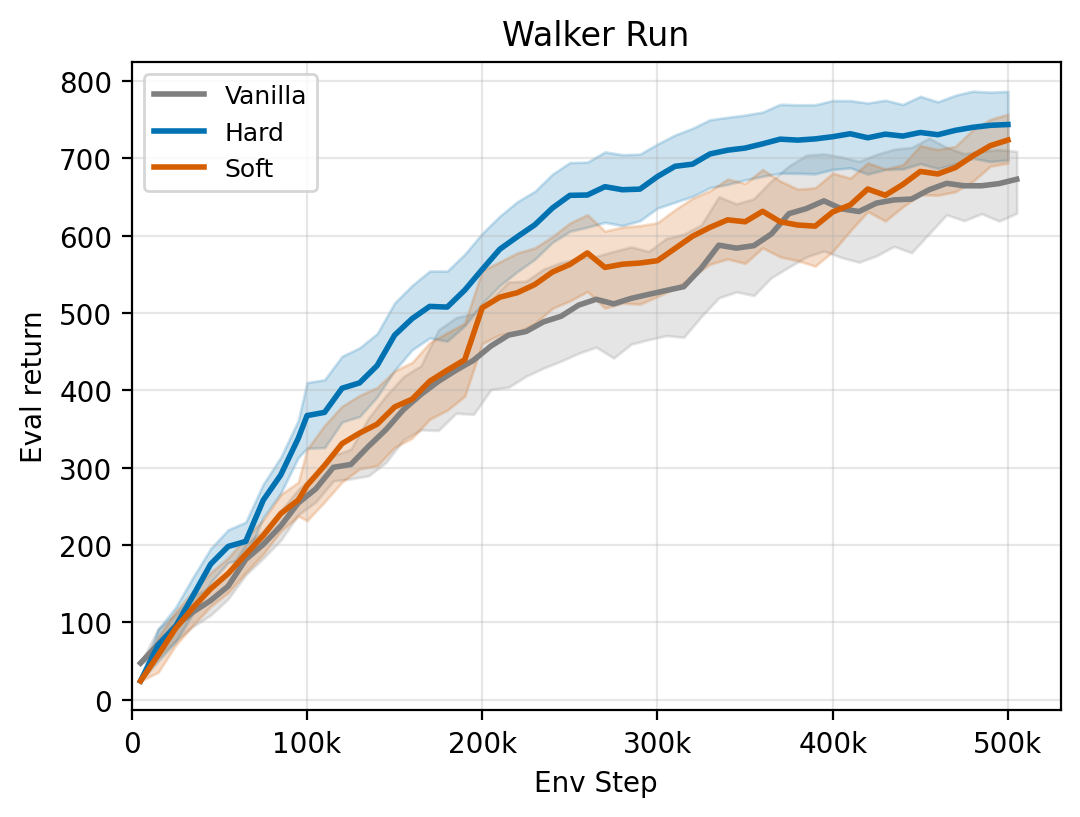}
\medskip
  \includegraphics[width=0.33\linewidth]{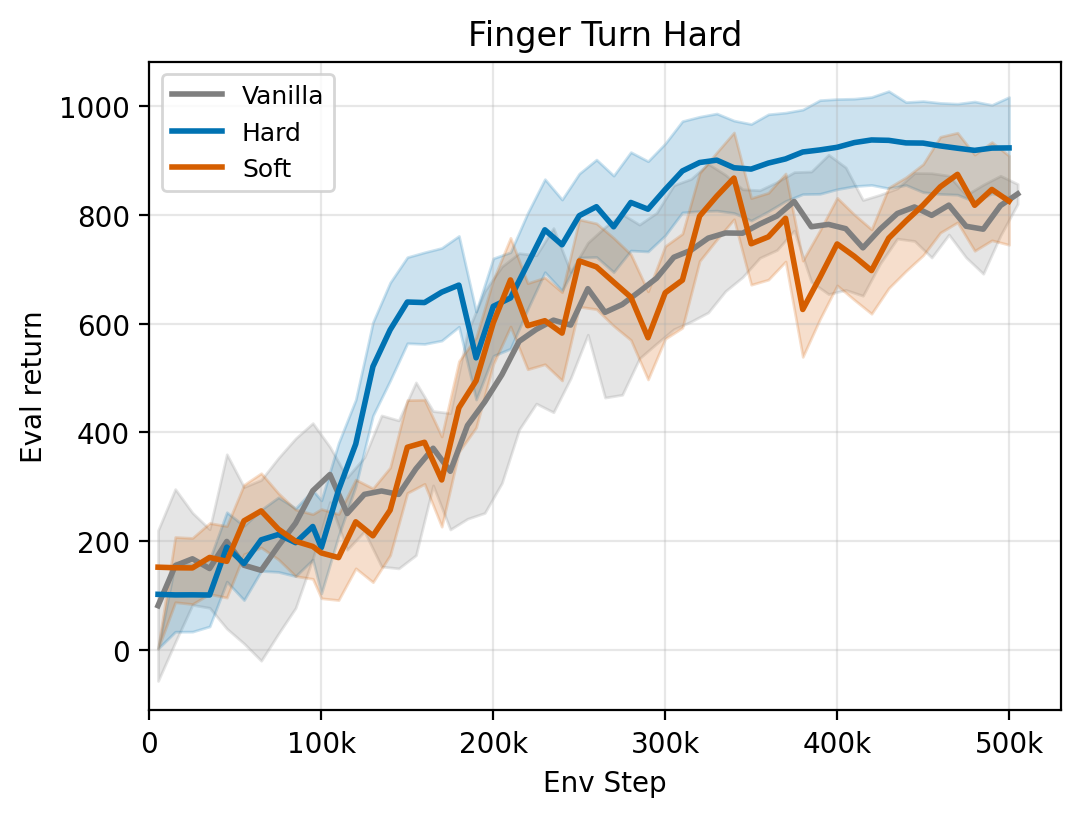}
  \includegraphics[width=0.33\linewidth]{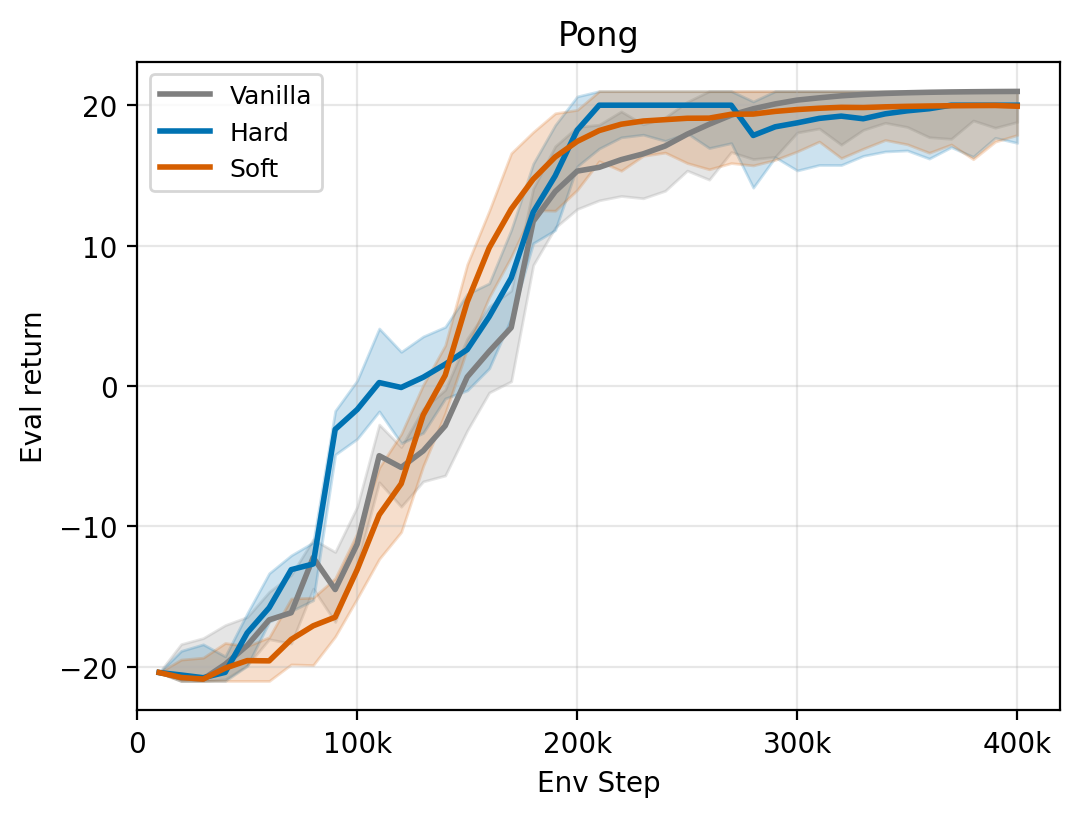}
\caption{Evaluation returns for vanilla DreamerV3 and density-guided hard and soft bootstrap boundaries across five tasks. Curves summarize 10 independent training seeds; shaded regions indicate one standard deviation across seeds. TensorBoard-style exponential smoothing with weight 0.7 is used for visualization.}
\label{fig:main_results}
\end{figure}

Figure~\ref{fig:main_results} compares vanilla DreamerV3 with density-guided hard and soft boundaries on four continuous-control tasks and Atari Pong. The curves indicate task-dependent benefits: some environments show faster learning with similar eventual returns, while others show a larger separation near the end of training. Neither boundary rule dominates across all environments.

\paragraph{Walker Walk.}
Both boundary variants reach high returns earlier than vanilla. Hard boundaries show the clearest advantage during the early-to-middle portion of training, while soft boundaries subsequently reach a similar performance level. The curves approach comparable returns by the end of the interaction budget. The principal benefit on this task is therefore earlier attainment of high performance rather than a substantial increase in final return.
\paragraph{Walker Run.}
Hard boundaries yield the highest mean return through most of the plotted training period. Soft boundaries improve more gradually and finish above vanilla, but below hard boundaries. Unlike Walker Walk, a visible gap remains at the final budget. The relative trajectories suggest that the discrete boundary rule is more effective than the selected soft-gate configuration on this task.
\paragraph{Cheetah Run.}
Both density-guided variants achieve higher late-training mean returns than vanilla, with soft boundaries finishing highest. Hard and soft exhibit different intermediate trajectories, including a temporary decline in the hard variant around the middle of training. Their relative ordering is therefore not constant over the full budget. The final separation motivates reporting both final-budget performance and a summary of learning efficiency, rather than selecting a single favorable intermediate checkpoint.
\paragraph{Finger Turn Hard.}
The hard-boundary variant achieves the strongest late-training performance. Its mean learning curve also changes less sharply during the later portion of training than the other displayed curves. Soft boundaries and vanilla are broadly comparable near the final budget, with crossings during training. The smoother hard mean curve should not be interpreted as a formal reduction in seed variance or policy-gradient variance; those quantities require separate measurements.
\paragraph{Pong.}
All three methods reach high positive returns. Hard boundaries make earlier progress over part of the initial learning period, while soft boundaries improve rapidly during a subsequent phase. Vanilla ultimately converges to the similar mean return near the end of training. Thus the Pong results indicate differences in learning dynamics rather than a consistent final-performance advantage for density-guided boundaries.

Taken together, these comparisons support evaluating boundary selection as a task-dependent modification to imagined learning. They do not establish that either hard or soft gating is uniformly preferable. In particular, the stronger late-training results on Cheetah Run and Finger Turn Hard coexist with near-convergent results on Walker Walk and on Pong.

To complement the learning curves, Table~\ref{tab:performance_stages}
summarizes evaluation returns at early, middle, and late stages
of training. These comparisons distinguish improvements during
learning from differences near the end of the training budget.
The stage definitions and extraction procedure are provided
in the table caption.


\begin{table}[t]
\centering
\small
\setlength{\tabcolsep}{5pt}
\caption{
Evaluation returns at early, middle, and late training stages.
Stage targets are 100k, 250k, and 500k environment steps for
DeepMind Control tasks, and 80k, 200k, and 400k for Pong.
Values are mean $\pm$ std.
}
\label{tab:performance_stages}
\begin{tabular}{llrrr}
\toprule
Environment & Method & Early & Middle & Late \\
\midrule
Walker Walk
& Vanilla & $544.2 \pm 128.0$ & $875.2 \pm 75.1$ & $945.6 \pm 26.9$ \\
& Hard    & $735.9 \pm 104.3$ & $917.3 \pm 69.0$ & $944.2 \pm 27.5$ \\
& Soft    & $663.5 \pm 82.6$  & $925.6 \pm 79.8$ & $951.3 \pm 24.2$ \\
\midrule
Walker Run
& Vanilla & $254.3 \pm 18.7$ & $495.6 \pm 61.5$ & $667.1 \pm 45.7$ \\
& Hard    & $367.2 \pm 45.3$ & $651.9 \pm 43.2$ & $743.5 \pm 46.6$ \\
& Soft    & $276.9 \pm 52.8$ & $562.6 \pm 54.7$ & $723.4 \pm 29.9$ \\
\midrule
Cheetah Run
& Vanilla & $382.0 \pm 73.6$ & $559.4 \pm 44.1$ & $653.1 \pm 57.2$ \\
& Hard    & $433.3 \pm 18.2$ & $629.5 \pm 45.8$ & $790.1 \pm 44.3$ \\
& Soft    & $433.3 \pm 18.7$ & $603.6 \pm 51.6$ & $815.6 \pm 30.1$ \\
\midrule
Finger Turn Hard
& Vanilla & $307.0 \pm 120.6$ & $605.7 \pm 89.3$ & $822.5 \pm 54.3$ \\
& Hard    & $185.4 \pm 89.0$  & $813.7 \pm 93.9$ & $923.7 \pm 93.6$ \\
& Soft    & $174.0 \pm 83.8$  & $738.2 \pm 81.9$ & $824.6 \pm 84.5$ \\
\midrule
Pong
& Vanilla & $-10.8 \pm 1.7$ & $16.9 \pm 3.5$ & $20.8 \pm 0.15$ \\
& Hard    & $-11.9 \pm 2.2$ & $19.5 \pm 1.58$ & $20.5 \pm 0.47$ \\
& Soft    & $-16.5 \pm 1.9$ & $19.0 \pm 2.11$ & $20.8 \pm 0.19$ \\
\bottomrule
\end{tabular}

\medskip
\begin{minipage}{0.98\linewidth}
\end{minipage}
\end{table}

The stage-wise comparisons indicate that the benefits depend
on both the task and the training stage. On Walker Walk, the
hard boundary achieves a higher early return, while the three
methods reach similar returns by late training. On Walker Run,
the hard variant has the highest reported mean at all three
stages. On Cheetah Run, both boundary variants exceed vanilla
at each reported stage, with the soft variant achieving the
highest late-stage mean. On Finger Turn Hard, the hard variant
initially trails vanilla but achieves higher middle- and
late-stage returns. Pong shows a different pattern: vanilla
has the highest early and late means, while the hard variant
leads at the middle stage. These descriptive comparisons
suggest task-dependent improvements rather than a uniform
advantage across all environments.

\subsection{Comparison with Fixed Bootstrap Boundaries}
\label{subsec:fixed_boundary_results}

We compare density-guided hard boundaries with fixed bootstrap
boundaries of six steps on Cheetah Run and five steps on Walker
Run to examine whether simple horizon shortening can account
for the observed performance differences. On Cheetah Run, the
fixed-boundary control retains the auxiliary displacement field,
score estimator, and density computations, while replacing
density-dependent boundary selection with a predetermined
six-step boundary. This provides a comparison of fixed and
density-selected boundaries with auxiliary components enabled
in both variants. On Walker Run, the fixed-boundary control
disables the auxiliary components and therefore compares
simple horizon shortening against the complete proposed method.
The Walker Run comparison does not isolate adaptive boundary
selection alone; the auxiliary-only ablation in the following
subsection provides a complementary assessment of the effect
of enabling density-guided boundaries.
\begin{figure}[t]
\centering
  \includegraphics[width=0.48\linewidth]{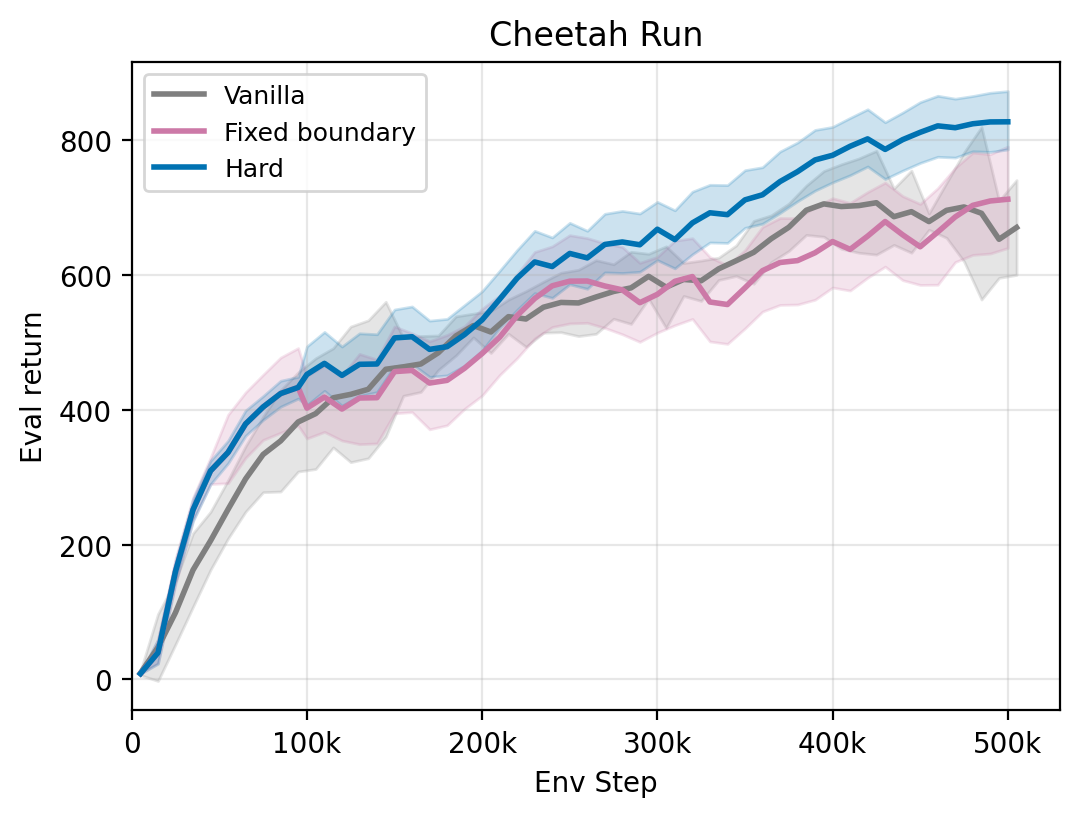}
  \includegraphics[width=0.48\linewidth]{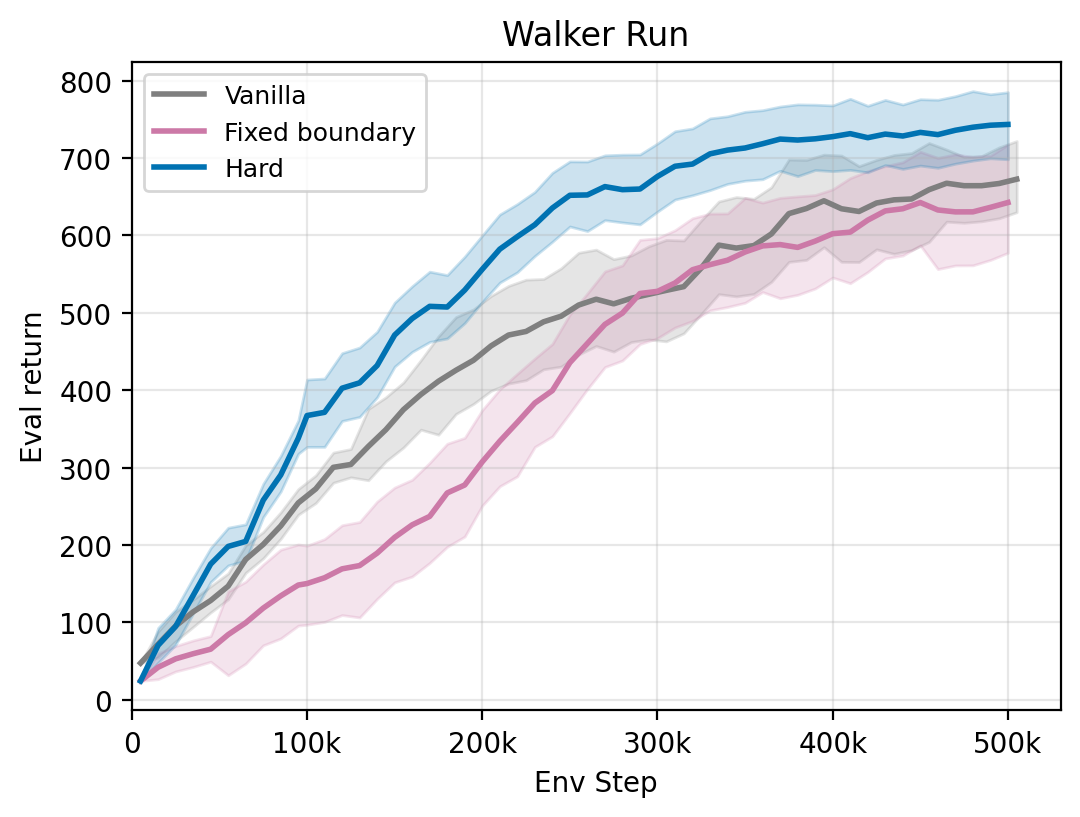}
\caption{Comparison of vanilla DreamerV3, fixed bootstrap boundaries, and density-guided hard boundaries on Cheetah Run (left)
and Walker Run (right). Fixed boundaries use six and five
steps, respectively. The Cheetah fixed control retains
the auxiliary components, whereas the Walker fixed control
disables them.}
\label{fig:fixed_boundary_results}
\end{figure}

\subsection{Auxiliary Mechanism and Boundary Selection}
\label{subsec:component_ablation_results}

The auxiliary-enabled, cutoff-off control separates the introduction of displacement and score learning from using their diagnostic to modify imagined returns. Comparing this control with vanilla measures the effect of the combined auxiliary machinery. Comparing this control with hard boundaries measures the
incremental effect of activating density-guided bootstrap
boundaries and their associated survival weights, provided
the remaining settings match. The supplied Cheetah Run curves show that the auxiliary-enabled cutoff-off agent itself improves over vanilla. The hard-boundary variant finishes above this control, although their ordering changes during training. On Walker Run, the hard variant leads the cutoff-off control during a substantial portion of training, while their final mean returns are similar. This pattern suggests that gating may improve intermediate
training performance even when the additional endpoint gain
is small. These results also limit the interpretation of a comparison against vanilla alone. Some observed improvement is already present when the auxiliary networks are enabled without boundary selection. Because the auxiliary losses use detached features, this observation does not directly establish that they improve the learned representation. Shared optimization details and stochastic computation can also distinguish the auxiliary-enabled control from vanilla.

To test whether the score correction changes the selected boundary,
we compared the standard hard rule with a drift-only rule on Cheetah
Run at the same threshold ($\tau=-900$). The drift-only proxy never crossed the threshold, so that agent retained
all 14 transitions at every logged point (Figure~\ref{fig:drift_boundary_behavior}). Over 400k--500k environment
steps, the score-corrected agent retained 6.35 transitions on average.
This change in boundary selection also translate into a higher
return 
: mean unsmoothed evaluation return over the same
window was 831.6 with score correction and 797.8 with drift alone.
Appendix~\ref{app:boundary_controls} reports the return curves and
examines the effects of boundary assignment and loss weighting.

\begin{figure}[t]
\centering
  \includegraphics[width=0.48\linewidth]{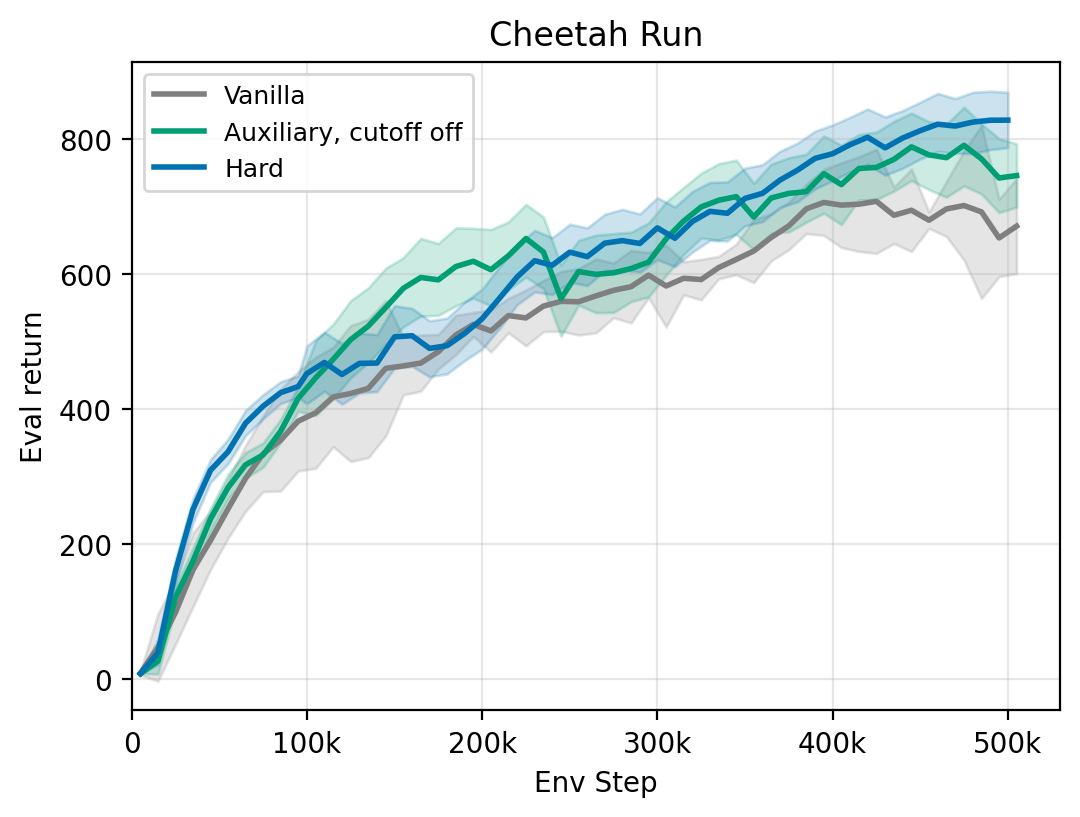}
  \includegraphics[width=0.48\linewidth]{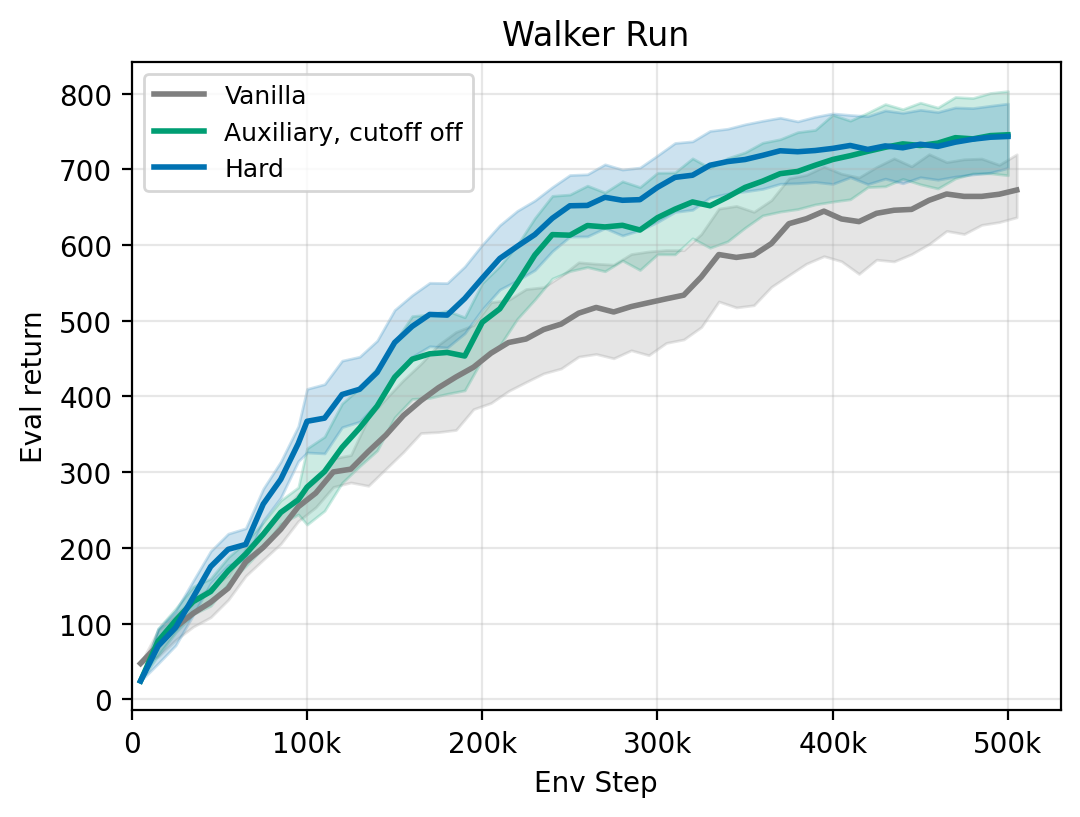}
  \caption{
Comparison of vanilla DreamerV3, the auxiliary-enabled
cutoff-off control, and density-guided hard boundaries on
Cheetah Run (left) and Walker Run (right). The cutoff-off
control retains auxiliary displacement and score learning
without using the density signal to modify imagined returns.
Curves use TensorBoard-style smoothing with coefficient 0.7.
}
\label{fig:auxiliary_ablation}
\end{figure}

\subsection{Sensitivity to the Density Threshold}
\label{subsec:threshold_sensitivity}

\begin{figure}[t]
\centering
  \includegraphics[width=\linewidth]{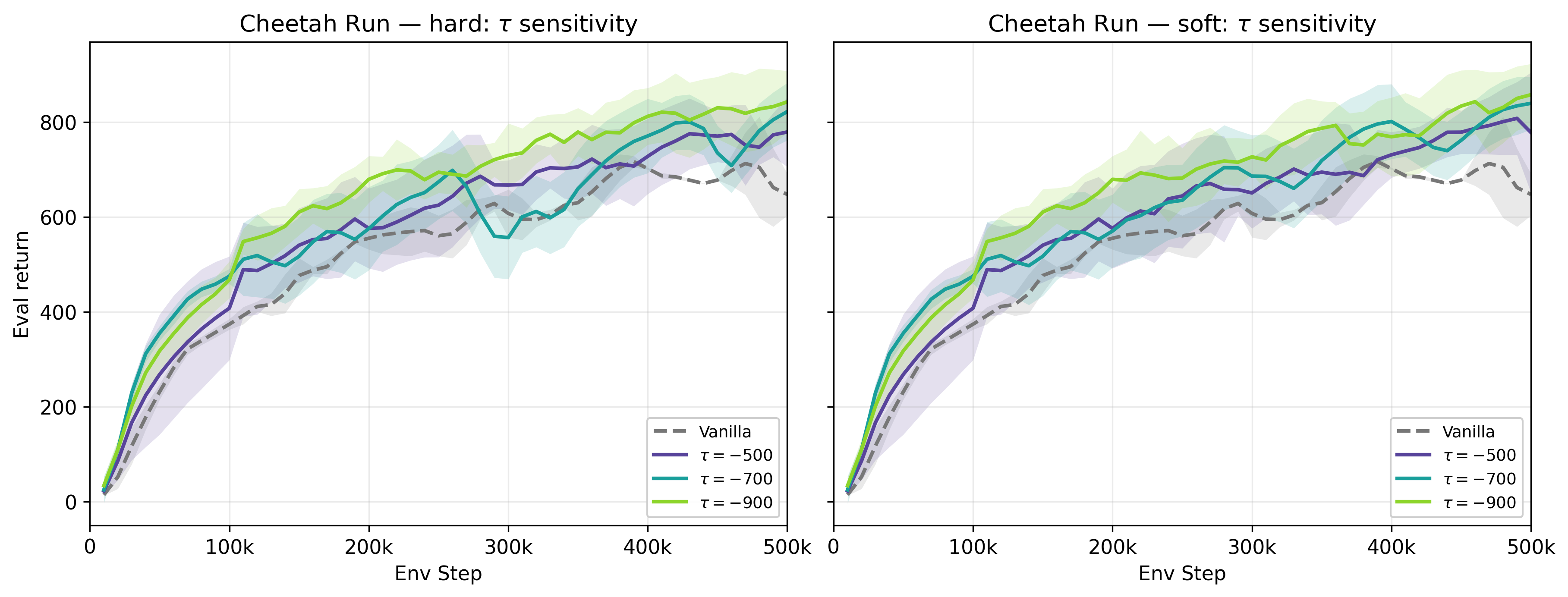}
\caption{
Density-threshold sensitivity on Cheetah Run for hard
boundaries (left) and soft boundaries (right), using
$\tau\in\{-500,-700,-900\}$. Each threshold configuration
uses 10 seeds, with boundary activation at 200k environment
steps, increment cap $C=200$, and minimum retained depth
$m=2$. Soft boundaries use temperature $T=100$.
The dashed curve shows the vanilla reference.
}
\label{fig:threshold_sensitivity}
\end{figure}

We examine sensitivity to the density threshold on Cheetah
Run using $\tau\in\{-500,-700,-900\}$ for both hard and soft
boundaries. The study fixes the soft temperature at $T=100$,
the increment cap at $C=200$, the minimum retained transition
count at $m=2$, and boundary activation at 200k logged
environment steps. 
Each threshold configuration uses 10 seeds. This study is therefore an exploratory sensitivity
analysis, separate from the 10-seed main comparisons. Figure~\ref{fig:threshold_sensitivity} shows that changing
the threshold affects learning after boundary activation.
The threshold variants follow similar trajectories during
warmup, when density-based selection is disabled, and
separate during subsequent training. In both hard and soft
modes, $\tau=-900$ achieves the highest return near the end
of the interaction budget, followed by $\tau=-700$ and
$\tau=-500$. The ordering is not maintained throughout
training, however, and the curves exhibit intermediate
crossings. All three threshold settings finish above the
displayed vanilla reference. For a fixed proxy sequence, decreasing the threshold makes
the boundary more permissive. In hard mode, fewer states
fail the threshold test; in soft mode, the continuation
gates increase. Consequently, a more negative threshold
produces no smaller survival mass when evaluated on the
same trajectory with the remaining settings fixed.
This ordering need not hold for effective horizons across
independently trained agents, because their policies,
latent representations, and proxy distributions can diverge. The late-training ordering in this experiment is consistent
with a benefit from a more permissive boundary setting
within the tested range. It does not establish that longer
effective horizons caused the improvement or that
$\tau=-900$ is generally optimal. 


\subsection{Behavior of the Density-Guided Boundaries}
\label{subsec:boundary_behavior}

\begin{figure}[tbp]
\centering
\begin{subfigure}[t]{0.32\linewidth}
    \centering
    \includegraphics[width=\linewidth]
        {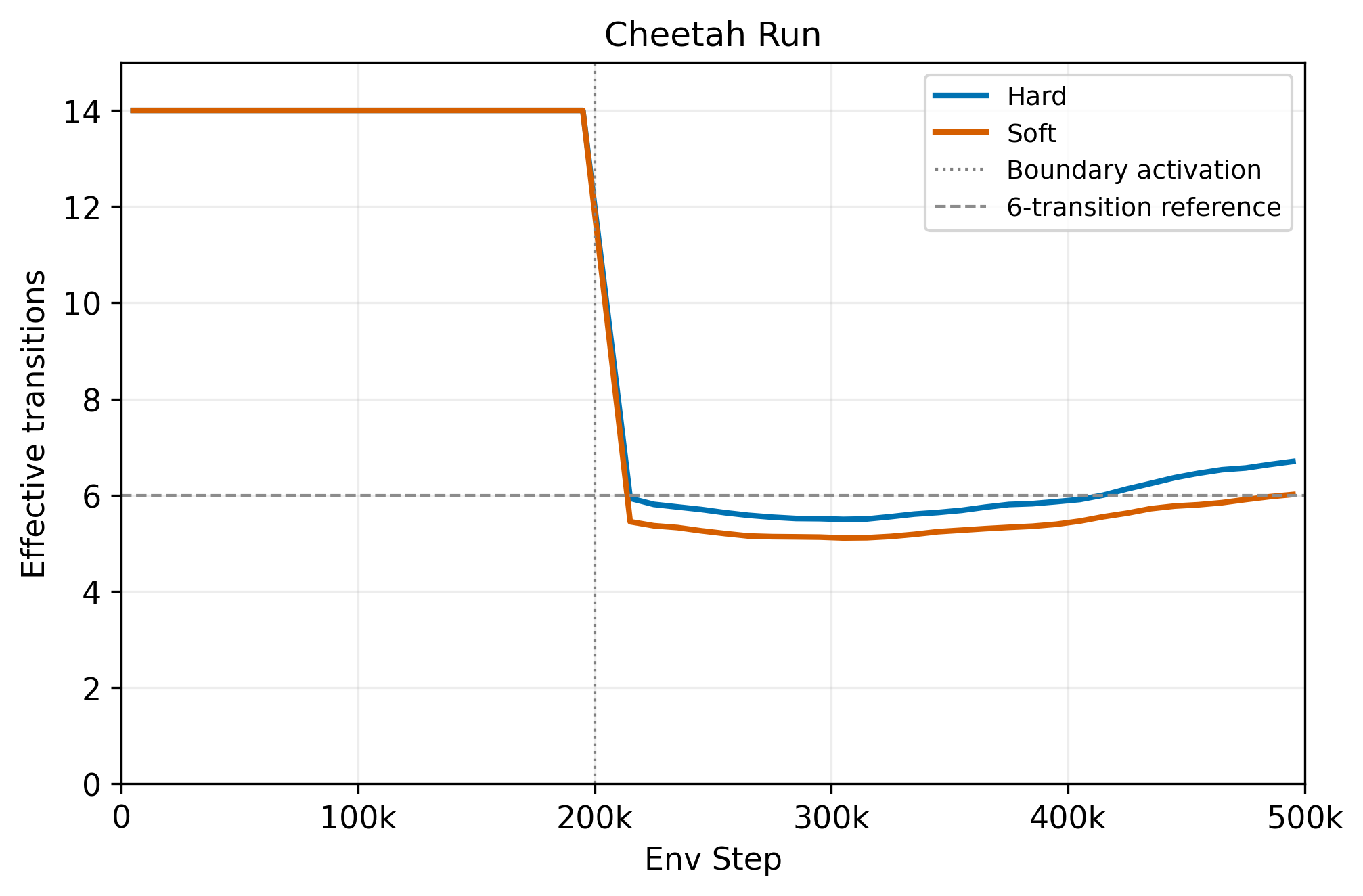}
    \caption{Cheetah: hard vs.\ soft}
    \label{fig:cheetah_boundary_behavior}
\end{subfigure}\hfill
\begin{subfigure}[t]{0.32\linewidth}
    \centering
    \includegraphics[width=\linewidth]
        {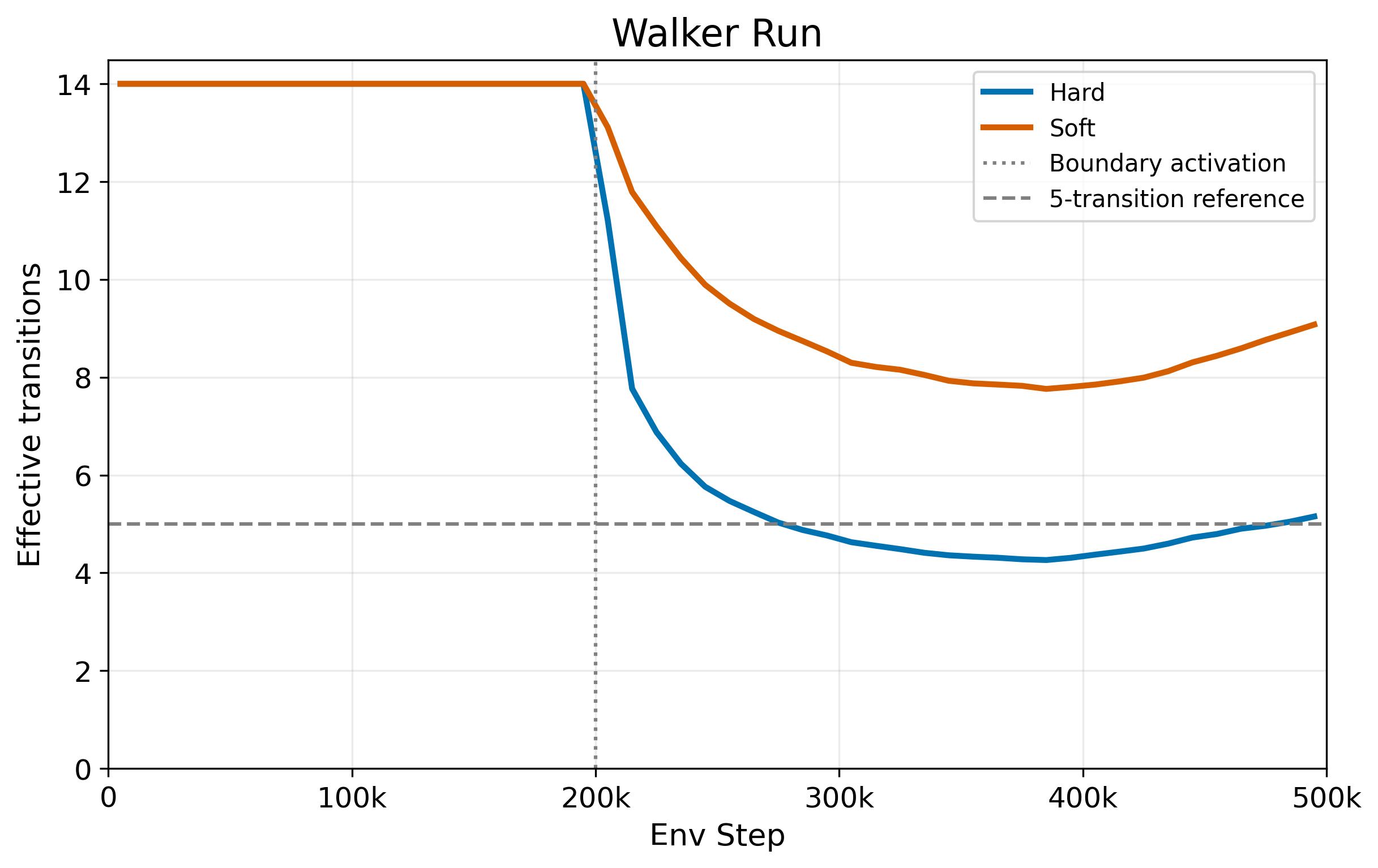}
    \caption{Walker: hard vs.\ soft}
    \label{fig:walker_boundary_behavior}
\end{subfigure}\hfill
\begin{subfigure}[t]{0.32\linewidth}
    \centering
    \includegraphics[width=\linewidth]
        {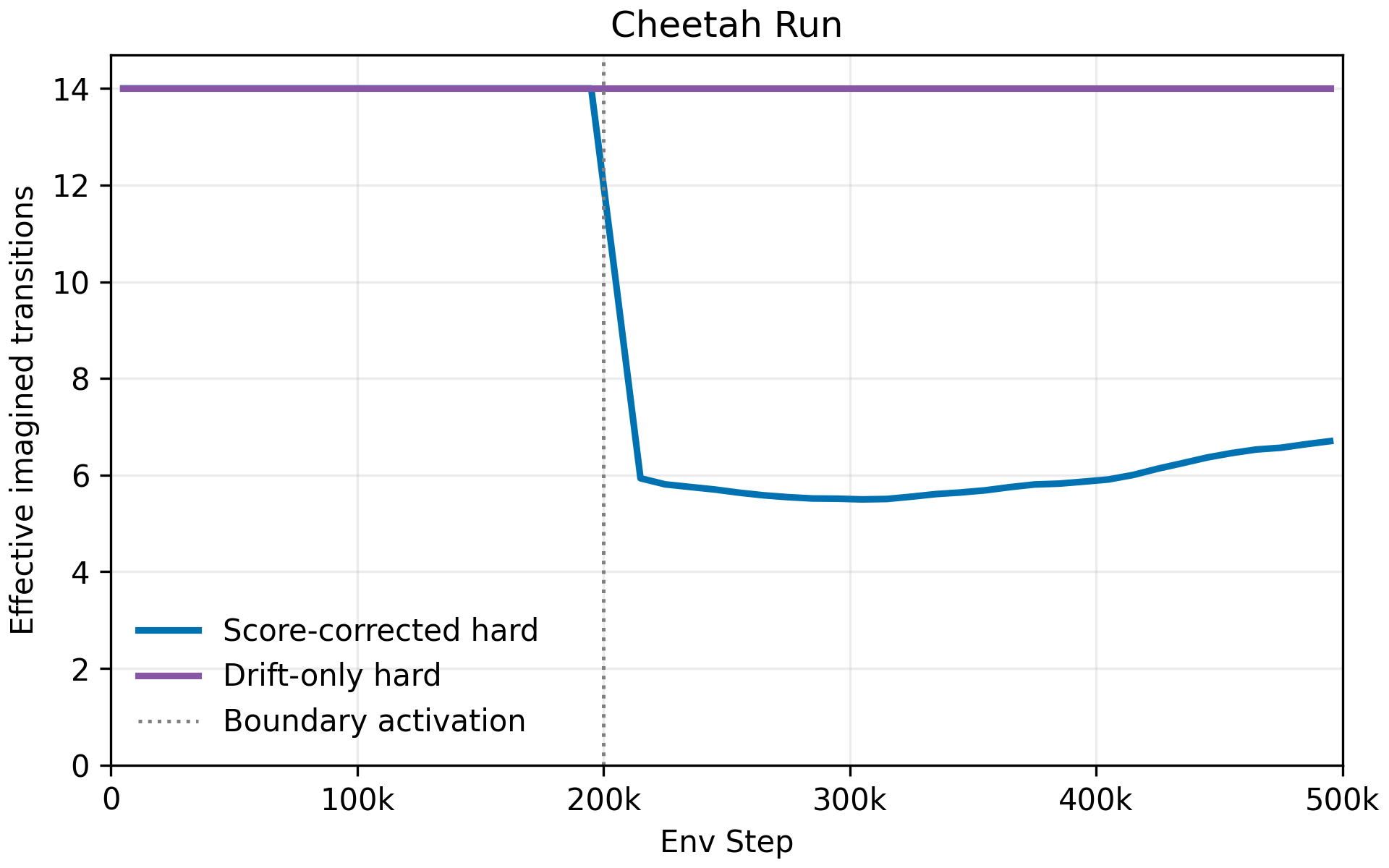}
    \caption{Cheetah: score correction}
    \label{fig:drift_boundary_behavior}
\end{subfigure}
\caption{Effective imagined transitions during training, with one run
per variant. Panels (a) and (b) compare hard and soft boundaries;
horizontal dashed lines mark the six- and five-transition references,
respectively. Panel (c) compares score-corrected and drift-only hard
boundaries at the same threshold. Vertical dotted lines
mark boundary activation at 200k environment steps.}
\label{fig:boundary_behavior}
\end{figure}

Figure~\ref{fig:boundary_behavior} shows how much of each imagined
trajectory contributes to learning. For hard boundaries, effective
length is the mean number of retained transitions; for soft
boundaries, it is the sum of survival weights averaged across
trajectories. All displayed variants retain the full effective
length of 14 during warmup.
After activation at 200k environment steps, the score-corrected hard
and soft variants both shorten their effective horizons. Their
ordering differs across tasks. On Cheetah Run, the soft variant
retains slightly less transition weight than the hard variant
(panel~a). On Walker Run, the soft variant retains substantially
more, ending near 9.1 effective transitions compared with 5.2 for
hard (panel~b). These curves come from separately trained agents,
so the differences reflect both the boundary rules and the
trajectories on which they act. The horizontal lines mark the fixed
depths used in Section~\ref{subsec:fixed_boundary_results}; they do
not imply that the adaptive horizons remain matched to those depths
throughout training. Panel~(c) shows why the choice of field matters for the hard rule:
at the same Cheetah threshold, the drift-only curve remains at 14
after activation, while the score-corrected curve drops.

Density-guided variants increased observed end-to-end runtime by approximately 2--7\% relative to vanilla DreamerV3. These measurements include evaluation and logging and do not isolate the computational cost of the density mechanism.
\section{Discussion and Limitations}
\label{sec:discussion}

\subsection{Adaptive Bootstrapping as a Model-Usage Decision}
\label{subsec:discussion_model_usage}

The central contribution of this work is to use a score-informed density diagnostic to decide how far an imagined return should expand before relying on the critic. This makes the use of model predictions trajectory-dependent while retaining stochastic latent dynamics. A boundary changes the source of the remaining value estimate: future model-generated rewards are replaced or attenuated in favor of a critic bootstrap. It therefore regulates model usage rather than eliminating estimation error.

The learning curves in Section~\ref{sec:results} suggest that the benefits of this decision depend on the task. On Walker Walk, the main difference is earlier attainment of high returns, with similar final performance. Cheetah Run and the hard-boundary variant on Finger Turn Hard show larger late-training improvements over vanilla. Pong exhibits differences in learning speed but no final-return advantage for the proposed variants. These patterns support distinguishing learning efficiency from final performance instead of treating either as a complete description of the method.

Hard and soft boundaries also make different choices about retaining future information. Hard gating removes downstream expansion after the first failed gate, whereas soft gating attenuates it cumulatively. The multiplicative soft survival weights can substantially reduce later contributions even when no individual gate is near zero. Consequently, a common threshold does not imply a common effective horizon or an equivalent intervention in learning. The different task-wise rankings of hard and soft variants motivate examining their induced survival profiles alongside return curves.

\subsection{Interpretation of the Density Proxy}
\label{subsec:discussion_proxy}

The quantity used for selection is an anchored accumulation of estimated divergence, not an exact latent likelihood. The auxiliary displacement field is fitted to posterior feature differences, while imagined trajectories are generated by a separate stochastic transition model. The score estimator describes a noise-smoothed replay-feature distribution, which need not coincide with the distribution visited during imagination. These differences prevent a direct identification of the implemented diagnostic with the density transported by a continuous-time flow.

Moreover, setting the initial proxy to zero means that the method tracks changes relative to each trajectory's starting point without estimating its absolute density. A trajectory starting in a poorly supported region can therefore initially receive the same proxy value as one starting in a well-supported region. Low density, density decrease, and prediction error are also distinct concepts. The observed usefulness of a boundary rule does not by itself establish that the underlying proxy is calibrated model uncertainty.

The score correction has a clear computational route to influencing learning: it changes the auxiliary field, its divergence, and potentially the gates. However, a large correction to the divergence is evidence of influence rather than evidence of accurate uncertainty estimation. Relating this signal to model prediction errors, critic errors, or return-estimation error remains an important direction for further analysis.

\subsection{Attribution of Performance Improvements}
\label{subsec:discussion_attribution}

Comparisons against vanilla measure the effect of the complete modification, including auxiliary computation and boundary selection. The auxiliary-enabled cutoff-off comparisons show why this distinction matters: part of the observed performance difference can be present without activating a boundary. Since the auxiliary losses use detached features, such differences should not automatically be interpreted as improved representation learning. Shared optimization details, gradient clipping, and changes in stochastic computation can also distinguish the training procedures.

Fixed-boundary controls address whether simple horizon shortening can explain the results. The evaluated fixed setting does not reproduce the hard variant's performance on the supplied Cheetah Run and Walker Run comparisons. Nevertheless, this does not establish superiority over every fixed horizon or scheduled horizon. Interpretation depends on matching activation times, auxiliary settings, and effective return depths. A fixed boundary matched only to the average adaptive horizon may still differ substantially in its distribution across trajectories and its evolution over training.

The Cheetah Run controls in Appendix~\ref{app:boundary_controls}
help distinguish changes in boundary behavior from changes in policy performance. At the same threshold ($\tau=-900$), the drift-only proxy never crossed the threshold, so that agent retained all 14 imagined transitions and coincides with an auxiliary-enabled agent without cutoff, while the score-corrected agent retained 6.35 on average over 400k--500k environment steps and achieved the higher mean evaluation return over that window. The score
correction therefore changed where the agent bootstrapped, and on these seeds that change
was accompanied by higher late-window return.


\subsection{Calibration and Nonstationarity}
\label{subsec:discussion_calibration}

The threshold, soft temperature, divergence cap, and activation time jointly determine boundary behavior. Their effects depend on the learned representation and auxiliary approximation. Proxy distributions can change as the replay buffer, world model, and policy evolve, so a threshold selected at one checkpoint need not maintain the same effective horizon later in training. Numerical values should therefore not be assumed to transfer unchanged between tasks or architectures.

Clipping limits the influence of extreme divergence estimates but changes the accumulated signal. When clipping is frequent, distinct estimates can produce identical increments, reducing the information available for selection. A low clipping rate avoids this particular saturation effect but does not establish that the remaining divergence estimates are accurate. Monitoring both clipping frequency and proxy distributions is therefore useful for interpreting a threshold sweep.

Warmup delays the use of the diagnostic while auxiliary learning proceeds. It does not guarantee that the score or displacement model has become accurate at activation, and abrupt activation can change targets and loss weights substantially. A gradual activation schedule or calibrated boundary criterion could be explored, but these are extensions rather than established features of the present method. 

\subsection{Computational and Implementation Scope}
\label{subsec:discussion_compute}

The implementation generates the full imagined trajectory before constructing the gates. A reduction in effective return depth therefore does not imply a reduction in trajectory-generation cost. Auxiliary training and spatial divergence estimation add computation, and any practical benefit must be assessed using measured training time or throughput as well as environment interactions. Comparisons of wall-clock cost require matched hardware and process concurrency.

Stochastic trace estimation avoids explicitly forming the full Jacobian, but it introduces another source of variability in the diagnostic. The required spatial derivatives include those of the score network. Improving score prediction alone does not necessarily improve these derivatives, and detaching the score input during divergence computation would remove the score's contribution entirely. Efficient and accurate derivative estimation is therefore a practical consideration for scaling the approach.

The formulation is compatible with imagined actor--critic learning beyond a single architecture, but the empirical evaluation here is confined to one DreamerV3 implementation. Results across continuous-control and visual discrete-action tasks demonstrate breadth within that implementation, not verified portability across world-model families. Applying the mechanism to another architecture would require checking the latent representation, auxiliary training distribution, return estimator, and gradient paths rather than transferring thresholds alone.

\subsection{Statistical and Theoretical Scope}
\label{subsec:discussion_evidence}

The main comparisons use 10 independent training seeds, with standard-deviation bands showing variation across runs. These bands are descriptive and do not directly quantify uncertainty in the difference between methods. Likewise, a smoother mean learning curve does not establish reduced estimator variance. Numerical performance comparisons should use unsmoothed per-seed summaries, with uncertainty procedures appropriate to the statistic being reported.

Variance across independently trained policies is distinct from the conditional variance of an imagined return or policy-gradient estimator at a fixed checkpoint. A theoretical guarantee about one quantity cannot be inferred from measurements of another. The relationship between the theory and experiments must therefore be stated in terms of the exact estimator, sources of randomness, and assumptions under consideration.

These limitations leave a focused empirical question: whether the score-informed diagnostic provides useful boundary decisions beyond simpler controls. The current results motivate that question and show that density-guided bootstrapping can change learning behavior while preserving stochastic imagination. Establishing when those changes improve estimation accuracy, and how reliably they transfer across tasks and implementations, requires the component controls and assumption-specific analysis described throughout the paper.

\section{Conclusion}
\label{sec:conclusion}

We presented density-guided bootstrap boundaries for latent world-model reinforcement learning, establishing a principled mechanism to govern model usage during policy learning. By evaluating spatial divergence along imagined latent paths via an auxiliary displacement field and a noise-conditioned score estimator, our framework constructs an anchored feature-space expansion diagnostic. Hard and soft continuation gates dynamically determine where return expansion transitions into value bootstrapping, preserving discounted boundary critic estimates and applying consistent survival weighting to actor and critic objectives.   

Theoretically, we established that the gated Bellman operator preserves strict contraction properties, characterized its local contraction dynamics under soft and hard gates, and proved that truncation monotonically bounds conditional target variance at the cost of absorbing critic approximation bias. Formally distinguishing this proxy from an exact path likelihood provides a clear boundary for theoretical claims and prevents misinterpreting spatial divergence as calibrated model uncertainty.   

Empirically, our DreamerV3 instantiation across DeepMind Control Suite and Atari Pong demonstrates that adaptive return horizons improve sample efficiency and final performance compared to vanilla DreamerV3 and fixed-horizon baselines. The performance trade-offs between discrete hard cutoffs and cumulative soft attenuation highlight task-dependent dynamics in return expansion. Controlled ablations confirm that gains stem from trajectory-dependent gate selection rather than horizon shortening or auxiliary loss scales alone.   

Future research will explore tighter couplings between feature-space divergence and local model prediction error, introduce adaptive thresholding mechanisms across evolving replay distributions, and extend the gating mechanism to broader model-based architectures and offline reinforcement learning settings.


\acks{The authors conceptualized, formulated, and provided the core mathematical ideas and directions for this work. Google Gemini was used during manuscript preparation to assist in drafting initial formal statements, theorems, and proofs in Section 6 and Appendices A-C. All derivations, theoretical formulations, empirical evaluations, and text were thoroughly reviewed, independently re-derived, verified, and finalized by the authors, who take full responsibility for the accuracy, claims, and integrity of all contents of this paper. PL is partially supported by NIH OT2 Award, 1OT2OD042825-01.}



\newpage
\bibliography{main}
\newpage
\appendix

\section{Overview}
This appendix provides rigorous mathematical proofs for the theoretical properties of density-guided bootstrap boundaries in latent world-model reinforcement learning. We formalize:
\begin{enumerate}
    \item Recursive Critic Bias Propagation (Theorem~\ref{thm:bias}).
    \item Strict Contraction of the Gated Evaluation Operator (Theorem~\ref{thm:contraction}).
    \item Conditional Return Variance Decomposition and Monotonicity (Theorem~\ref{thm:variance}, Appendix~B).
    \item Score-Divergence Dependent Contraction Rate and Hard Discontinuity (Lemma~\ref{lem:local_rate} and Corollary~\ref{cor:hard_disc}, Appendix~C).
    \item DreamerV3 Implementation Details
    \item Additional Controls for Boundary Selection
(Appendix~\ref{app:boundary_controls}):
Cheetah Run experiments examining score correction, assignment of
boundaries to trajectories, and normalization of the surviving loss weights.
\end{enumerate}

\subsection{Proof of Theorem~\ref{thm:bias} (Recursive Critic Bias Propagation)}

\begin{proof}
Recall the recursive definition of the gated return target:
\begin{equation}
G_t^g = \hat{r}_{t+1} + d_{t+1}\big[(1 - \lambda g_{t+1})V_\psi(x_{t+1}) + \lambda g_{t+1}G_{t+1}^g\big],
\end{equation}
where $d_{t+1} = \gamma\hat{c}_{t+1} \in [0, \gamma]$, $\lambda \in [0, 1]$, and $g_{t+1} = g(l_{t+1}) \in [0, 1]$. Using the identity $1 = (1 - \lambda g_{t+1}) + \lambda g_{t+1}$, we expand the Bellman equation $V^\pi(x_t) = \mathbb{E}[\hat{r}_{t+1} + d_{t+1}V^\pi(x_{t+1}) \mid \bar{x}_t]$ as
\begin{equation}
V^\pi(x_t) = \mathbb{E}\!\big[\hat{r}_{t+1} + d_{t+1}(1 - \lambda g_{t+1})V^\pi(x_{t+1}) + d_{t+1}\lambda g_{t+1}V^\pi(x_{t+1}) \;\big|\; \bar{x}_t\big].
\end{equation}
Substituting $V_\psi(x_{t+1}) = V^\pi(x_{t+1}) + e(x_{t+1})$ into $\mathbb{E}[G_t^g \mid \bar{x}_t]$ and subtracting $V^\pi(x_t)$ cancels $\hat{r}_{t+1}$ and $d_{t+1}(1 - \lambda g_{t+1})V^\pi(x_{t+1})$, yielding
\begin{equation}
\mathrm{Bias}(G_t^g \mid \bar{x}_t) = \mathbb{E}\!\big[d_{t+1}(1 - \lambda g_{t+1})e(x_{t+1}) \;\big|\; \bar{x}_t\big] + \mathbb{E}\!\big[d_{t+1}\lambda g_{t+1}(G_{t+1}^g - V^\pi(x_{t+1})) \;\big|\; \bar{x}_t\big].
\end{equation}
Let $\mathcal{F}_{t+1} = \sigma(\bar{x}_t, a_t, \xi_t, x_{t+1})$. Since $d_{t+1}$ and $g_{t+1}$ are $\mathcal{F}_{t+1}$-measurable and $G_{t+1}^g$ depends on $\mathcal{F}_{t+1}$ only through the augmented Markov state $\bar{x}_{t+1} = (x_{t+1}, l_{t+1})$, the Law of Iterated Expectations gives
\begin{align}
\mathbb{E}\!\big[d_{t+1}\lambda g_{t+1}(G_{t+1}^g - V^\pi(x_{t+1})) \;\big|\; \bar{x}_t\big] &= \mathbb{E}\!\Big[d_{t+1}\lambda g_{t+1}\,\mathbb{E}\big[G_{t+1}^g - V^\pi(x_{t+1}) \;\big|\; \bar{x}_{t+1}\big] \;\Big|\; \bar{x}_t\Big] \nonumber \\
&= \mathbb{E}\!\big[d_{t+1}\lambda g_{t+1}\mathrm{Bias}(G_{t+1}^g \mid \bar{x}_{t+1}) \;\big|\; \bar{x}_t\big],
\end{align}
establishing~\eqref{eq:bias_recursion}.
\end{proof}

\subsection{Convergence of the Gated Evaluation Operator}

\begin{theo}[Strict Contraction of the Gated Evaluation Operator]\label{thm:contraction}
Let $\mathcal{S} = \mathcal{X} \times \mathbb{R}$ be the augmented state space, $\gamma \in [0, 1)$ the discount factor, $d(x,a) = \gamma \hat{c}(x,a) \in [0, \gamma]$ the continuation-discount factor, $\lambda \in [0, 1]$, and $g : \mathcal{S} \to [0, 1]$ a measurable continuation gate. On the Banach space $\mathcal{B}(\mathcal{S})$ of bounded measurable real-valued functions equipped with the supremum norm $\|V\|_\infty = \sup_{\bar{x} \in \mathcal{S}} |V(\bar{x})|$, define the gated $\lambda$-return Bellman operator $T_g^\lambda : \mathcal{B}(\mathcal{S}) \to \mathcal{B}(\mathcal{S})$ implicitly by
\begin{equation}\label{eq:gated_bellman}
(T_g^\lambda V)(\bar{x}) = \mathbb{E}_{a \sim \pi(\cdot \mid x),\, \bar{x}' \sim p_{\phi,v}(\cdot \mid \bar{x}, a)}\!\Big[\hat{r}(x,a) + d(x,a)\big((1 - \lambda g(\bar{x}'))V(\bar{x}') + \lambda g(\bar{x}')(T_g^\lambda V)(\bar{x}')\big)\Big],
\end{equation}
assuming bounded predicted rewards $|\hat{r}(x,a)| \le R_{\max}$. Then $T_g^\lambda$ is well-defined and is a strict $\gamma_g$-contraction mapping under $\|\cdot\|_\infty$ with contraction modulus
\begin{equation}\label{eq:contraction_modulus}
\gamma_g \triangleq \sup_{x, a, \bar{x}'} \frac{d(x,a)(1 - \lambda g(\bar{x}'))}{1 - d(x,a)\lambda g(\bar{x}')} \le \gamma < 1,
\end{equation}
and hence admits a unique fixed point $V^{\pi, g} \in \mathcal{B}(\mathcal{S})$.
\end{theo}

\begin{rem}[Finite-Horizon Non-Stationary Recursion]
In the finite-horizon implementation (Section~5.5) with step-dependent conditioning $u_t = t/H$, extending the augmented state space to $\mathcal{S}_H = \mathcal{X} \times \mathbb{R} \times \{0, \dots, H\}$ with terminal boundary condition $V(\cdot, \cdot, H) = V_\psi(\cdot)$ yields the identical step-wise contraction modulus $\gamma_g \le \gamma$.
\end{rem}

\begin{proof}
First, for any fixed $V \in \mathcal{B}(\mathcal{S})$, consider the affine operator $\mathcal{K}_V : \mathcal{B}(\mathcal{S}) \to \mathcal{B}(\mathcal{S})$ defined by $(\mathcal{K}_V W)(\bar{x}) = \mathbb{E}_{a,\bar{x}'}[\hat{r}(x,a) + d(x,a)(1-\lambda g(\bar{x}'))V(\bar{x}') + d(x,a)\lambda g(\bar{x}')W(\bar{x}')]$. Because $0 \le d(x,a)\lambda g(\bar{x}') \le \gamma\lambda < 1$, $\mathcal{K}_V$ is a contraction on $\mathcal{B}(\mathcal{S})$ with modulus at most $\gamma\lambda$, so its unique fixed point $W^* = T_g^\lambda V \in \mathcal{B}(\mathcal{S})$ is well-defined.

Next, for $d \in [0, \gamma]$ and $g \in [0, 1]$, define $f(d,g) \triangleq \frac{d(1 - \lambda g)}{1 - d\lambda g}$. Its partial derivative with respect to $g$ satisfies
\begin{equation}
\frac{\partial f}{\partial g} = \frac{-d\lambda(1 - d\lambda g) - d(1 - \lambda g)(-d\lambda)}{(1 - d\lambda g)^2} = \frac{-d\lambda(1 - d)}{(1 - d\lambda g)^2} \le 0.
\end{equation}
Hence $f(d,g)$ attains its maximum over $g \in [0,1]$ at $g = 0$, giving $f(d,g) \le f(d,0) = d \le \gamma < 1$. Taking the supremum over $(x, a, \bar{x}')$ confirms $\gamma_g \le \gamma < 1$.

Now let $V_1, V_2 \in \mathcal{B}(\mathcal{S})$, and denote $A \triangleq \|V_1 - V_2\|_\infty$ and $B \triangleq \|T_g^\lambda V_1 - T_g^\lambda V_2\|_\infty$. Write $\alpha(x,a,\bar{x}') \triangleq d(x,a)(1 - \lambda g(\bar{x}')) \ge 0$ and $\beta(x,a,\bar{x}') \triangleq d(x,a)\lambda g(\bar{x}') \in [0, \gamma\lambda]$. For any $\bar{x} \in \mathcal{S}$, Jensen's inequality gives
\begin{align}
\big|(T_g^\lambda V_1)(\bar{x}) - (T_g^\lambda V_2)(\bar{x})\big| &\le \mathbb{E}_{a, \bar{x}'}\!\Big[\alpha(x,a,\bar{x}')\big|V_1(\bar{x}') - V_2(\bar{x}')\big| + \beta(x,a,\bar{x}')\big|(T_g^\lambda V_1)(\bar{x}') - (T_g^\lambda V_2)(\bar{x}')\big|\Big] \nonumber \\
&\le \mathbb{E}_{a, \bar{x}'}\!\big[\alpha(x,a,\bar{x}') A + \beta(x,a,\bar{x}') B\big].
\end{align}
By definition of $\gamma_g$, we have $\frac{\alpha(x,a,\bar{x}')}{1 - \beta(x,a,\bar{x}')} \le \gamma_g$, and thus $\alpha(x,a,\bar{x}') \le \gamma_g(1 - \beta(x,a,\bar{x}'))$ pointwise for all $(x,a,\bar{x}')$. Substituting this bound inside the expectation yields
\begin{equation}
\alpha(x,a,\bar{x}') A + \beta(x,a,\bar{x}') B \le \gamma_g(1 - \beta(x,a,\bar{x}'))A + \beta(x,a,\bar{x}') B = \gamma_g A + \beta(x,a,\bar{x}')(B - \gamma_g A).
\end{equation}
We consider two exhaustive cases:
\begin{enumerate}
    \item If $B \le \gamma_g A$, then $\|T_g^\lambda V_1 - T_g^\lambda V_2\|_\infty \le \gamma_g \|V_1 - V_2\|_\infty$ holds immediately.
    \item If $B > \gamma_g A$, then $B - \gamma_g A > 0$. Because $\beta(x,a,\bar{x}') \le \gamma\lambda < 1$ uniformly over $(x,a,\bar{x}')$, we have $\beta(x,a,\bar{x}')(B - \gamma_g A) \le \gamma\lambda(B - \gamma_g A)$. Taking expectations over $(a, \bar{x}')$ and then taking the supremum over $\bar{x} \in \mathcal{S}$ on the left-hand side gives
    \begin{equation}
    B \le \gamma_g A + \gamma\lambda(B - \gamma_g A) \implies (1 - \gamma\lambda)B \le (1 - \gamma\lambda)\gamma_g A.
    \end{equation}
    Since $1 - \gamma\lambda \ge 1 - \gamma > 0$, dividing both sides by $1 - \gamma\lambda$ yields $B \le \gamma_g A$.
\end{enumerate}
In both cases, $\|T_g^\lambda V_1 - T_g^\lambda V_2\|_\infty \le \gamma_g \|V_1 - V_2\|_\infty$ with $\gamma_g \le \gamma < 1$. By Banach's Fixed-Point Theorem, $T_g^\lambda$ converges uniquely to a fixed point $V^{\pi, g} \in \mathcal{B}(\mathcal{S})$.
\end{proof}

\section{Conditional Return Variance Decomposition and Monotonicity}

\begin{theo}[Conditional Return Variance Decomposition and Monotonicity]\label{thm:variance}
Let $\bar{x}_t = (x_t, l_t)$ be the augmented state at step $t$, let $g : \mathcal{S} \to [0, 1]$ be a state-dependent continuation gate $g_{k+1} = g(\bar{x}_{k+1})$, and define the critic's multi-step Bellman residual $e_k^g(\bar{x}_k) \triangleq V_\psi(x_k) - \mathbb{E}[G_k^g \mid \bar{x}_k]$.
\begin{enumerate}
    \item \textbf{(Exact Variance Decomposition)} The conditional target variance decomposes into a one-step bootstrap/residual variance and a non-negative downstream rollout variance $\mathcal{V}_{\mathrm{roll}}(g)$:
    \begin{equation}\label{eq:var_decomp}
    \mathrm{Var}(G_t^g \mid \bar{x}_t) = \mathrm{Var}\!\Big(\hat{r}_{t+1} + d_{t+1}V_\psi(x_{t+1}) - d_{t+1}\lambda g_{t+1} e_{t+1}^g(\bar{x}_{t+1}) \;\Big|\; \bar{x}_t\Big) + \mathcal{V}_{\mathrm{roll}}(g),
    \end{equation}
    where
    \begin{equation}
    \mathcal{V}_{\mathrm{roll}}(g) \triangleq \sum_{k=t+1}^{H-1} \mathbb{E}\!\left[\left(\prod_{j=t+1}^{k} (d_j \lambda g_j)^2\right) A_k^g(\bar{x}_k) \;\middle|\; \bar{x}_t\right] \ge 0, \qquad A_k^g(\bar{x}_k) \triangleq \mathrm{Var}\!\big(\mathbb{E}[G_k^g \mid \mathcal{F}_{k+1}] \;\big|\; \bar{x}_k\big).
    \end{equation}
    In particular, a hard boundary at step $t+1$ ($g_{t+1} = 0$) completely eliminates downstream rollout stochasticity ($\mathcal{V}_{\mathrm{roll}} = 0$), yielding $\mathrm{Var}(G_t^{g_{t+1}=0} \mid \bar{x}_t) = \mathrm{Var}(\hat{r}_{t+1} + d_{t+1}V_\psi(x_{t+1}) \mid \bar{x}_t)$.

    \item \textbf{(Monotonicity under Vanishing Critic Bellman Residual)} Suppose the critic is at its gated fixed point so that $e_k^g(\bar{x}_k) = 0$ for all $k \ge t+1$ (e.g., $V_\psi(x) = V^{\pi, g}(x)$). Then $A_k^g(\bar{x}_k) = \mathrm{Var}(\hat{r}_{k+1} + d_{k+1}V_\psi(x_{k+1}) \mid \bar{x}_k)$ is independent of $g$, and for any two gate functions satisfying $0 \le g^{(1)}(\bar{x}) \le g^{(2)}(\bar{x}) \le 1$ pointwise on $\mathcal{S}$, total conditional variance is monotonically non-decreasing:
    \begin{equation}\label{eq:var_mono}
    \mathrm{Var}(G_t^{g^{(1)}} \mid \bar{x}_t) \le \mathrm{Var}(G_t^{g^{(2)}} \mid \bar{x}_t).
    \end{equation}
\end{enumerate}
\end{theo}

\begin{proof}
Let $\mathcal{F}_{k+1} \triangleq \sigma(\bar{x}_k, a_k, \xi_k, x_{k+1})$. Because $\hat{r}_{k+1}$, $d_{k+1}$, $x_{k+1}$, and $g_{k+1} = g(\bar{x}_{k+1})$ are $\mathcal{F}_{k+1}$-measurable, and $G_{k+1}^g$ depends on $\mathcal{F}_{k+1}$ solely through the augmented Markov state $\bar{x}_{k+1}$, we have
\begin{align}
\mathbb{E}[G_k^g \mid \mathcal{F}_{k+1}] &= \hat{r}_{k+1} + d_{k+1}\big[(1 - \lambda g_{k+1})V_\psi(x_{k+1}) + \lambda g_{k+1}\mathbb{E}[G_{k+1}^g \mid \bar{x}_{k+1}]\big] \nonumber \\
&= \hat{r}_{k+1} + d_{k+1}V_\psi(x_{k+1}) - d_{k+1}\lambda g_{k+1} e_{k+1}^g(\bar{x}_{k+1}), \\
\mathrm{Var}(G_k^g \mid \mathcal{F}_{k+1}) &= (d_{k+1}\lambda g_{k+1})^2 \mathrm{Var}(G_{k+1}^g \mid \bar{x}_{k+1}).
\end{align}
By the Law of Total Variance conditioned on $\bar{x}_k$,
\begin{align}
\mathrm{Var}(G_k^g \mid \bar{x}_k) &= \mathrm{Var}\!\big(\mathbb{E}[G_k^g \mid \mathcal{F}_{k+1}] \;\big|\; \bar{x}_k\big) + \mathbb{E}\!\big[\mathrm{Var}(G_k^g \mid \mathcal{F}_{k+1}) \;\big|\; \bar{x}_k\big] \nonumber \\
&= A_k^g(\bar{x}_k) + \mathbb{E}\!\big[(d_{k+1}\lambda g_{k+1})^2 \mathrm{Var}(G_{k+1}^g \mid \bar{x}_{k+1}) \;\big|\; \bar{x}_k\big].
\end{align}
Unrolling this recursion from $k = t$ to $H - 1$ and using the terminal condition $\mathrm{Var}(G_H^g \mid \bar{x}_H) = \mathrm{Var}(V_\psi(x_H) \mid \bar{x}_H) = 0$ establishes~\eqref{eq:var_decomp}. Setting $g_{t+1} = 0$ sets the common factor $(d_{t+1}\lambda g_{t+1})^2 = 0$ in every summand of $\mathcal{V}_{\mathrm{roll}}(g)$, proving Part~1.

For Part~2, when $e_k^g \equiv 0$, we have $\mathbb{E}[G_k^g \mid \mathcal{F}_{k+1}] = \hat{r}_{k+1} + d_{k+1}V_\psi(x_{k+1})$, so $A_k^g(\bar{x}_k) = A_k(\bar{x}_k) \ge 0$ is independent of $g$. Because $0 \le g^{(1)}(\bar{x}_j) \le g^{(2)}(\bar{x}_j) \le 1$ implies $0 \le (d_j \lambda g_j^{(1)})^2 \le (d_j \lambda g_j^{(2)})^2$ pointwise along every trajectory, each non-negative summand in $\mathcal{V}_{\mathrm{roll}}(g^{(1)})$ is bounded above by the corresponding summand in $\mathcal{V}_{\mathrm{roll}}(g^{(2)})$, yielding~\eqref{eq:var_mono}.
\end{proof}

\section{Score-Divergence Dependent Contraction Rate}

\begin{lem}[Score-Divergence Dependent Contraction Rate]\label{lem:local_rate}
Let $l_{t+1} = \sum_{j=0}^{t} \delta_j$ be the accumulated negative score-divergence proxy along an imagined trajectory prefix, where $\delta_j = \mathrm{clip}(-\hat{D}_j, -C, C)$, and let $g(l) = \big[1 + \exp(-(l - \tau)/T)\big]^{-1}$ denote the soft continuation gate with temperature $T > 0$ and threshold $\tau$. In the non-terminating regime $\hat{c}(x,a) = 1$ (so $d(x,a) = \gamma$), the state-dependent effective contraction factor $\gamma_g(l_{t+1})$ of the gated Bellman operator $T_g^\lambda$ is given by
\begin{equation}\label{eq:local_contraction}
\gamma_g(l_{t+1}) = \frac{\gamma\big(1 + e^{-(l_{t+1} - \tau)/T} - \lambda\big)}{1 + e^{-(l_{t+1} - \tau)/T} - \gamma\lambda}.
\end{equation}
Furthermore:
\begin{enumerate}
    \item The effective contraction factor is strictly monotonically decreasing in $l_{t+1}$:
    \begin{equation}
    \frac{\partial \gamma_g}{\partial l_{t+1}} < 0 \qquad \text{for all } \lambda \in (0, 1),\; \gamma \in (0, 1).
    \end{equation}
    \item The asymptotic limits satisfy
    \begin{equation}
    \lim_{l_{t+1} \to -\infty} \gamma_g(l_{t+1}) = \gamma, \qquad \lim_{l_{t+1} \to +\infty} \gamma_g(l_{t+1}) = \frac{\gamma(1 - \lambda)}{1 - \gamma\lambda} < \gamma.
    \end{equation}
\end{enumerate}
\end{lem}

\begin{proof}
Substituting $g_{t+1} = \big(1 + e^{-(l_{t+1} - \tau)/T}\big)^{-1}$ and $d = \gamma$ into the local contraction ratio $\frac{d(1 - \lambda g_{t+1})}{1 - d\lambda g_{t+1}}$ from Theorem~\ref{thm:contraction} gives
\begin{equation}
\gamma_g(l_{t+1}) = \frac{\gamma\left(1 - \frac{\lambda}{1 + e^{-(l_{t+1} - \tau)/T}}\right)}{1 - \frac{\gamma\lambda}{1 + e^{-(l_{t+1} - \tau)/T}}} = \frac{\gamma\big(1 + e^{-(l_{t+1} - \tau)/T} - \lambda\big)}{1 + e^{-(l_{t+1} - \tau)/T} - \gamma\lambda}.
\end{equation}
Define $u(l_{t+1}) \triangleq e^{-(l_{t+1} - \tau)/T} > 0$, noting that $\frac{du}{dl_{t+1}} = -\frac{1}{T}u < 0$. Differentiating $\gamma_g$ with respect to $u$ yields
\begin{equation}
\frac{\partial \gamma_g}{\partial u} = \gamma \cdot \frac{\partial}{\partial u}\!\left(\frac{u + 1 - \lambda}{u + 1 - \gamma\lambda}\right) = \gamma \cdot \frac{(u + 1 - \gamma\lambda) - (u + 1 - \lambda)}{(u + 1 - \gamma\lambda)^2} = \frac{\gamma\lambda(1 - \gamma)}{(u + 1 - \gamma\lambda)^2} > 0.
\end{equation}
By the chain rule,
\begin{equation}
\frac{\partial \gamma_g}{\partial l_{t+1}} = \frac{\partial \gamma_g}{\partial u} \cdot \frac{du}{dl_{t+1}} = \left(\frac{\gamma\lambda(1 - \gamma)}{(u + 1 - \gamma\lambda)^2}\right)\!\left(-\frac{1}{T}u\right) < 0.
\end{equation}
Finally, as $l_{t+1} \to -\infty$, we have $u \to +\infty$ and $\lim_{u \to +\infty} \frac{u + 1 - \lambda}{u + 1 - \gamma\lambda} = 1$, so $\gamma_g(l_{t+1}) \to \gamma$. As $l_{t+1} \to +\infty$, we have $u \to 0^+$ and $\gamma_g(l_{t+1}) \to \frac{\gamma(1 - \lambda)}{1 - \gamma\lambda} < \gamma$.
\end{proof}

\begin{coro}[Hard Boundary Contraction Discontinuity]\label{cor:hard_disc}
For the hard continuation gate $g^{\mathrm{hard}}(l_{t+1}) = \mathbb{I}(l_{t+1} \ge \tau)$ in the non-terminating regime $\hat{c}(x,a) = 1$, the local contraction factor exhibits a step-function phase transition at threshold $\tau$:
\begin{equation}
\gamma_g^{\mathrm{hard}}(l_{t+1}) = \begin{cases}
\dfrac{\gamma(1 - \lambda)}{1 - \gamma\lambda}, & \text{if } l_{t+1} \ge \tau, \\[8pt]
\gamma, & \text{if } l_{t+1} < \tau.
\end{cases}
\end{equation}
When the accumulated negative score-divergence drops below $\tau$, $T_g^\lambda$ abruptly collapses to a one-step Bellman evaluation operator with contraction factor $\gamma$, eliminating downstream rollout predictions.
\end{coro}

\section{DreamerV3 Implementation Details}
\label{app:dreamer_implementation}

\begin{algorithm}[t]
\caption{Dreamer V3 World-Model and Auxiliary Update}
\label{alg:wm_update}
\begin{algorithmic}[1]
\REQUIRE Replay batch of observations, actions, rewards, continuation labels, and reset indicators; world model parameters $\phi$; displacement parameters $\omega$; noise predictor parameters $\eta$.
\STATE Encode observations and infer posterior RSSM sequence using observed actions and reset indicators.
\STATE Compute base Dreamer V3 loss $\mathcal{L}_{\text{WM}}$ (reconstruction, reward, continuation, dynamics, representation).
\STATE Extract detached posterior features $\bar{x}_i \gets \text{sg}(F_\phi(z_i))$.
\STATE Align each action $b_i$ with transition $\bar{x}_i \to \bar{x}_{i+1}$ and construct mask $m_i$ excluding transitions across episode resets.
\STATE Set $u_i \gets i / (L - 1)$ for replay sequence length $L$.
\STATE Compute masked displacement loss:
\begin{equation*}
\mathcal{L}_{\text{disp}} \gets \frac{\sum_i m_i \|f_\omega(u_i, \bar{x}_i, b_i) - (\bar{x}_{i+1} - \bar{x}_i)\|_2^2}{d \max(1, \sum_i m_i)}.
\end{equation*}
\STATE Sample noise scales $\alpha \sim \mathcal{U}[\alpha_{\min}, \alpha_{\max}]$ and independent Gaussian noise $\epsilon \sim \mathcal{N}(0, \mathbf{I}_d)$.
\STATE Compute unweighted noise prediction loss:
\begin{equation*}
\mathcal{L}_{\text{score}} \gets \text{mean}\left(\frac{1}{d} \|\epsilon_\eta(\bar{x} + \alpha \epsilon, \alpha) - \epsilon\|_2^2\right).
\end{equation*}
\STATE Form $\mathcal{L}_{\text{model}} \gets \mathcal{L}_{\text{WM}} + \beta_{\text{disp}} \mathcal{L}_{\text{disp}} + \beta_{\text{score}} \mathcal{L}_{\text{score}}$.
\STATE Update $(\phi, \omega, \eta)$ using shared model optimizer and global gradient clipping.
\RETURN Detached posterior states as starting points for imagined behavior learning.
\end{algorithmic}
\end{algorithm}

\begin{algorithm}[t]
\caption{Dreamer V3 Imagined Behavior Update with Density Gates}
\label{alg:behavior_update}
\begin{algorithmic}[1]
\REQUIRE Detached posterior starting states; actor $\pi_\theta$; critic $V_\psi$; trained world model and auxiliary networks; boundary settings $(\tau, T, C, m, N_{\text{warm}})$.
\STATE Hold world model and auxiliary parameters fixed during behavior update.
\STATE Generate $H+1$ feature entries using standard RSSM imagination and actions sampled from $\pi_\theta$.
\STATE Predict rewards $\hat{r}_t$, continuation probabilities $\hat{c}_t$, and values $V_\psi(x_t)$.
\STATE Set $l_0 \gets 0$.
\FOR{$t = 0, \dots, H-1$}
    \STATE Set $u_t \gets t / H$ and evaluate $v_t(x) = f_\omega(u_t, x, a_t) + \frac{\sigma}{2} \epsilon_\eta(x, \sigma)$.
    \STATE Estimate $\hat{D}_t \gets \xi_t^\top [\nabla_x v_t(x)]_{x=x_t} \xi_t$ using independent Rademacher probe $\xi_t$, holding $u_t$ and $a_t$ fixed.
    \STATE Set $l_{t+1} \gets l_t + \text{clip}(-\hat{D}_t, -C, C)$.
\ENDFOR
\STATE Detach $l_{0:H}$ after computing spatial derivatives.
\STATE Construct hard or soft gates $g_t$ (Section 5.4), applying warmup and minimum depth overrides.
\STATE Compute detached survival weights $S_t \gets \prod_{j=0}^t g_j$.
\STATE Set $G_H \gets V_\psi(x_H)$.
\FOR{$t = H-1, \dots, 0$}
    \STATE $G_t \gets \hat{r}_{t+1} + \gamma \hat{c}_{t+1} [(1 - \lambda g_{t+1}) V_\psi(x_{t+1}) + \lambda g_{t+1} G_{t+1}]$.
\ENDFOR
\STATE Construct inherited detached discount weights $W_0 = 1$ and $W_t = \prod_{j=0}^{t-1} \gamma \hat{c}_j$.
\STATE Set $\Omega_t \gets W_t S_t$ for $t = 0, \dots, H-1$.
\STATE Update return normalization statistics using mode-dependent target selection.
\STATE Update actor using base dynamics-gradient or likelihood-ratio surrogate weighted by $\Omega_t$, with entropy weighted by $S_t$.
\STATE Update distributional critic using detached targets $G_t$, detached features, and weights $\Omega_t$.
\STATE Average both objectives over full batch-time tensor ($B \times H$) without normalizing by surviving mass.
\STATE Apply base slow-critic update schedule.
\end{algorithmic}
\end{algorithm}

\paragraph{Indexing and normalization.}
The implementation setting \texttt{imag\_horizon}=15
produces 15 feature entries and 14 return-training
transitions, corresponding to $H=14$ in these algorithms.
Replay and imagination use different sequence-position
normalizations; neither is an ODE integration step size.
The implemented return-normalization statistics exclude
zero-survival targets in hard and fixed modes, whereas
soft mode retains all targets.

\paragraph{Gradient handling.}
Auxiliary feature inputs are detached during model training.
During divergence estimation, the spatial derivative includes
the score network's dependence on its feature input.
The resulting density proxies, gates, and survival weights
are detached from actor--critic optimization. Freezing the
world-model parameters during dynamics-based actor learning
does not remove the differentiable paths through imagined
states and actions.

\paragraph{Computational scope.}
The full imagined rollout is generated before applying
the gates. The method changes return expansion and
optimization weights; it does not implement early
termination of RSSM trajectory generation.

\section{Additional Controls for Boundary Selection}
\label{app:boundary_controls}

\begin{figure}[tbp]
\centering
\begin{subfigure}[t]{0.32\linewidth}
    \centering
    \includegraphics[width=\linewidth]
        {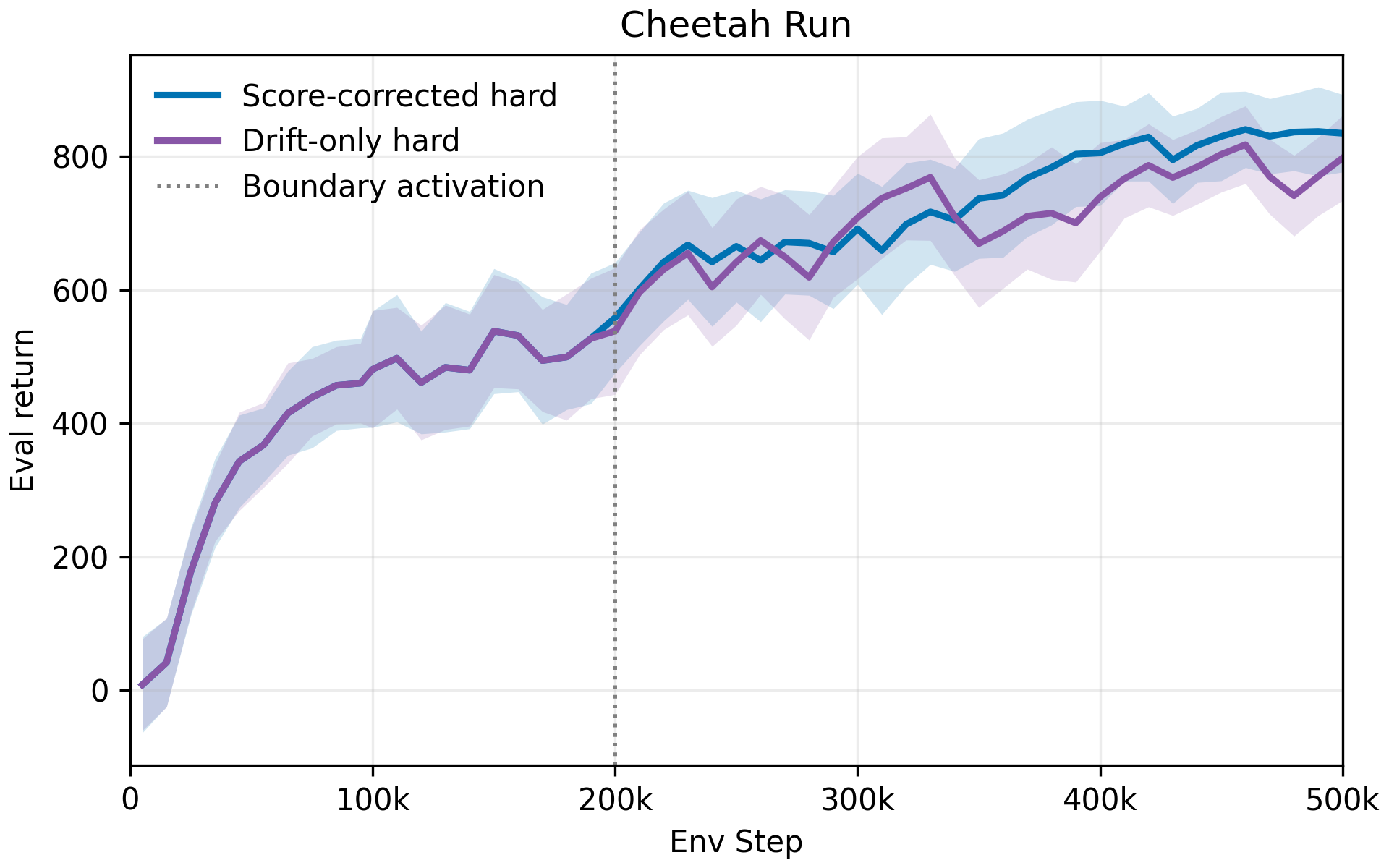}
    \caption{Score correction}
    \label{fig:app_drift_return}
\end{subfigure}\hfill
\begin{subfigure}[t]{0.32\linewidth}
    \centering
    \includegraphics[width=\linewidth]
        {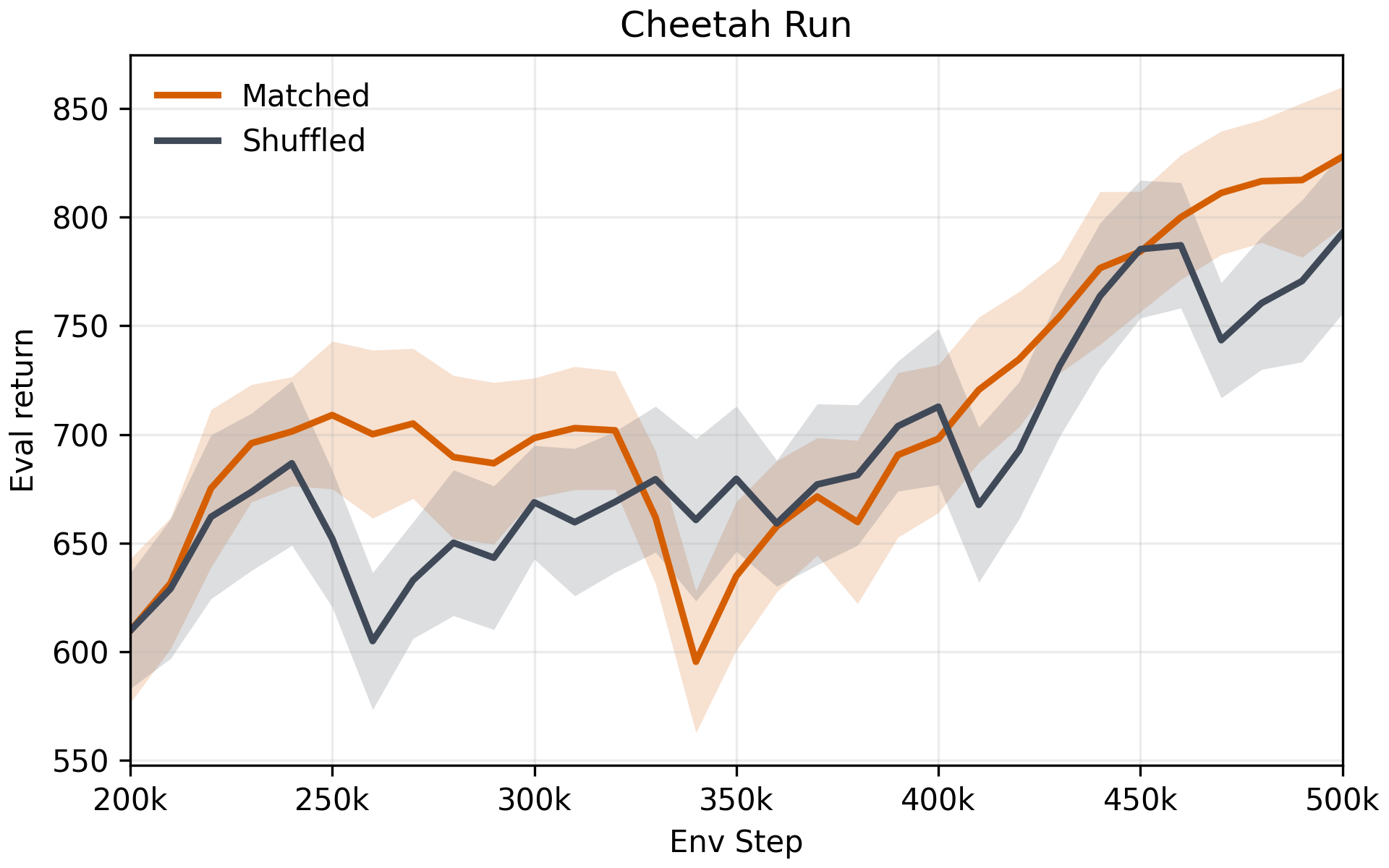}
    \caption{Boundary assignment}
    \label{fig:app_shuffle_return}
\end{subfigure}\hfill
\begin{subfigure}[t]{0.32\linewidth}
    \centering
    \includegraphics[width=\linewidth]
        {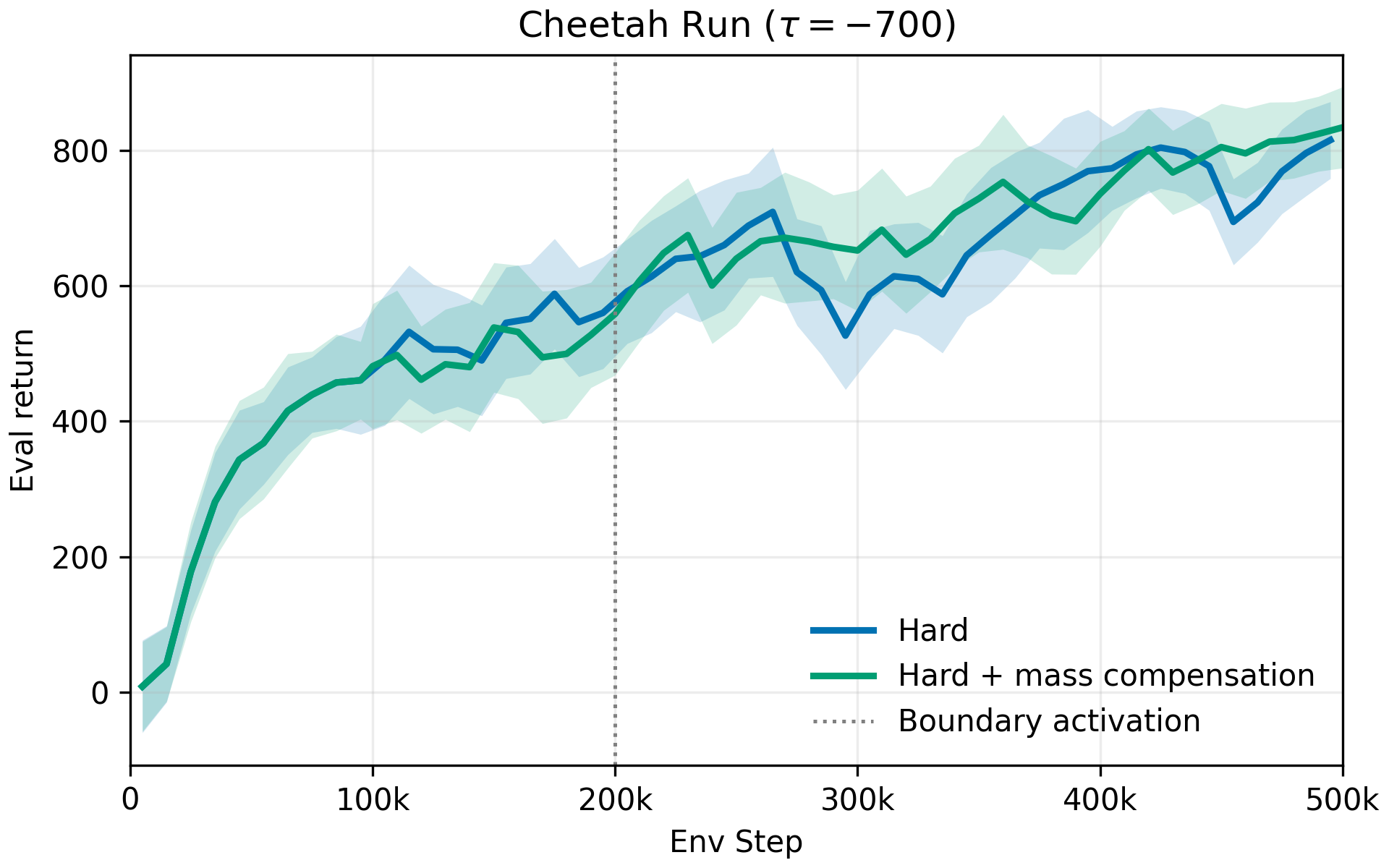}
    \caption{Loss normalization}
    \label{fig:app_mass_return}
\end{subfigure}
\caption{Evaluation return for three hard-boundary controls on
Cheetah Run. Panel~(b) begins at the shared checkpoint from which
the matched and shuffled runs branch. Return curves use exponential
smoothing with weight 0.7 for display. The numerical comparisons
below use unsmoothed evaluations from 400,000 to 500,000 environment
steps.}
\label{fig:cheetah_boundary_controls}
\end{figure}

Figure~\ref{fig:cheetah_boundary_controls} reports three Cheetah Run comparisons using 10 seeds and boundary activation at
200,000 environment steps. Panel (a) uses $\tau=-900$, panel (b) uses $\tau=-700$,
and panel (c) uses $\tau=-700$. Each panel changes
a different part of the standard hard-boundary method.

\paragraph{Score correction.}
Panel~(a) shows the evaluation returns corresponding to the effective
horizons in Figure~\ref{fig:drift_boundary_behavior}. Mean late-window return was 797.8 with drift alone and 831.6 with score correction.
At the same threshold, the drift-only agent retained all 14 imagined
transitions, whereas the score-corrected agent retained 6.35 on
average over this window. 
The correction therefore changed where the agent bootstrapped, and it did improve late-window return on these
seeds. This comparison holds the threshold fixed; it does not compare
the two signals at equal retained depth.

\paragraph{Boundary assignment.}
In panel~(b), both runs start from the same 200,000-step checkpoint.
The matched variant applies each trajectory's own boundary. The
shuffled variant permutes complete boundary sequences across
trajectories within each batch, preserving the set of boundaries
while changing which trajectory receives each one. The return curves
cross during training, and the matched run finishes higher on the
smoothed curve. 

\paragraph{Loss normalization.}
In panel~(c), both agents use the score-corrected hard boundary.
The standard update averages over the full batch of imagined
transitions, so cutting off more transitions reduces the total loss
weight. The mass-compensated update rescales the surviving weights
to offset that reduction. Mean late-window return was 824.2 with
compensation and 784.8 without it. In these runs, restoring the loss
weight did not reduce late-window performance. 
 
\vskip 0.2in

\end{document}